\documentclass[10pt,journal,compsoc]{IEEEtran}
\ifCLASSOPTIONcompsoc
  \usepackage[nocompress]{cite}
\else
  \usepackage{cite}
\fi

\ifCLASSINFOpdf
\else
\fi
\usepackage{graphicx}
\usepackage{overpic}
\usepackage[colorlinks,linkcolor=blue]{hyperref}
\usepackage{amsmath}
\usepackage{multirow}

\usepackage{array}
\newcolumntype{C}[1]{>{\centering\let\newline\\\arraybackslash\hspace{0pt}}m{#1}}
\usepackage{amssymb}
\usepackage{amsfonts}
\usepackage{algorithmic}
\usepackage{xcolor}
\usepackage{booktabs}
\usepackage{bbm}
\usepackage{bm}
\usepackage{color, colortbl}
\newcommand{\ie}{\textit{i}.\textit{e}.}
\newcommand{\eg}{\textit{e}.\textit{g}.}
\newcommand{\etal}{\textit{et} \textit{al}.}

\ifCLASSOPTIONcompsoc
 \usepackage[caption=false,font=footnotesize,labelfont=sf,textfont=sf]{subfig}
\else
  \usepackage[caption=false,font=footnotesize]{subfig}
\fi
\begin{document}
\title{Image Inpainting with Edge-guided Learnable Bidirectional Attention Maps}
%
%
%
%
	
\author{Dongsheng~Wang*,
	    Chaohao~Xie*,
        Shaohui~Liu, 
        Zhenxing Niu,
        Wangmeng~Zuo,~\IEEEmembership{Senior Member,~IEEE}
    \IEEEcompsocitemizethanks{\IEEEcompsocthanksitem D. Wang and C. Xie are with the School
	of Computer Science and Technology, Harbin Institute of Technology, Harbin, China, e-mail: (wds02467@163.com; viousxie@outlook.com).*The two authors contribute equally.
	\IEEEcompsocthanksitem W. Zuo and S.~Liu are with the School
	of Computer Science and Technology, Harbin Institute of Technology, Harbin, 150001, China and are also with Peng Cheng Lab, Shenzhen, China, e-mail: (cswmzuo@gmail.com, shliu@hit.edu.cn).
\IEEEcompsocthanksitem Z. Niu is with the DAMO Academy at Alibaba Group, Hangzhou, China, e-mail: (zhenxing.nzx@alibaba-inc.com).}
\thanks{Manuscript received xxx; revised xxx (Corresponding author: Wangmeng Zuo)}}

%
%

\markboth{The IEEE Transactions on Pattern Analysis and Machine Intelligence }%
{Shell \MakeLowercase{\textit{et al.}}: Bare Advanced Demo of IEEEtran.cls for IEEE Computer Society Journals}
\IEEEtitleabstractindextext{%
\begin{abstract}
	For image inpainting, the convolutional neural networks (CNN) in previous methods often adopt standard convolutional operator, which treats valid pixels and holes indistinguishably. As a result, they are limited in handling irregular holes and tend to produce color-discrepant and blurry inpainting result.
	Partial convolution (PConv) copes with this issue by conducting masked convolution and feature re-normalization conditioned only on valid pixels, but the mask-updating is handcrafted and independent with image structural information.
	In this paper, we present an edge-guided learnable bidirectional attention map (Edge-LBAM) for improving image inpainting of irregular holes with several distinct merits.
	Instead of using a hard 0-1 mask, a learnable attention map module is introduced for learning feature re-normalization and mask-updating in an end-to-end manner.
	Learnable reverse attention maps are further proposed in the decoder for emphasizing on filling in unknown pixels instead of reconstructing all pixels.
	Motivated by that the filling-in order is crucial to inpainting results and largely depends on image structures in exemplar-based methods,
	%
	{we further suggest a multi-scale edge completion network to predict coherent edges. Our Edge-LBAM method contains dual procedures, including structure-aware mask-updating guided by predict edges and attention maps generated by masks for feature re-normalization.}
	Extensive experiments show that our Edge-LBAM is effective in generating coherent image structures and preventing color discrepancy and blurriness, and performs favorably against the state-of-the-art methods in terms of qualitative metrics and visual quality.
The source code and pre-trained models are available at~\url{https://github.com/wds1998/Edge-LBAM}.  
\end{abstract}

}

\maketitle
\thispagestyle{plain}
\begin{figure*}[hbt]
	\small
	\setlength{\tabcolsep}{2.0pt}
	\centering
		\begin{tabular}{cccc}
		\includegraphics[width=.234\textwidth]{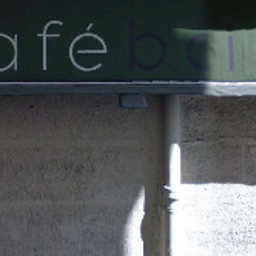}  &
		\includegraphics[width=.234\textwidth]{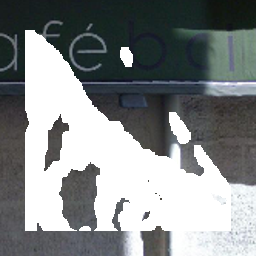}  &
		\includegraphics[width=.234\textwidth]{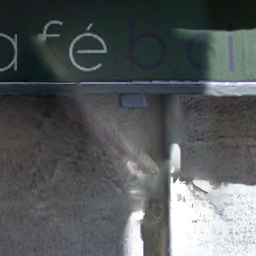}  &
		\includegraphics[width=.234\textwidth]{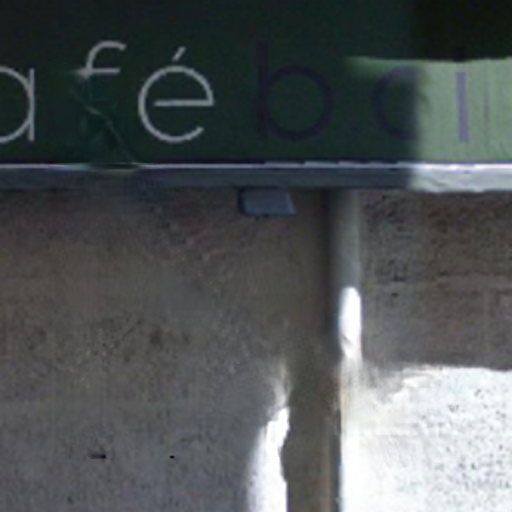} \\
		\scriptsize{(a) Original} & \scriptsize{(b) Input} &  \scriptsize{(c) GL~\cite{IizukaGL}} & \scriptsize{(d) PConv~\cite{partialconv2017}}\\
		\vspace{-2mm}\\
		\includegraphics[width=.234\textwidth]{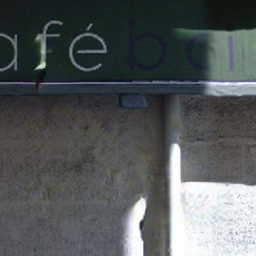}  &
		\includegraphics[width=.234\textwidth]{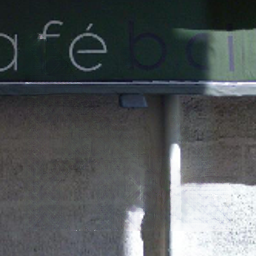}  &
		\includegraphics[width=.234\textwidth]{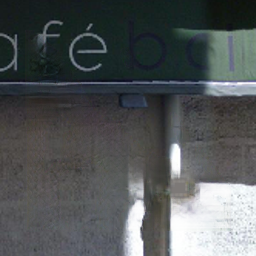}  &
		\includegraphics[width=.234\textwidth]{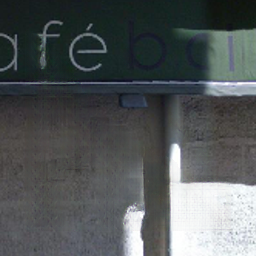}  \\
		
		\scriptsize{(e) DFv2~\cite{yu2018free}} & \scriptsize{(f) EC~\cite{nazeri2019edgeconnect}} & \scriptsize{(g) MEDFE~\cite{liu2020rethinking}} & \scriptsize{(h) Ours} \\
		\vspace{-2mm}\\	
	\end{tabular}

	\caption{Qualitative comparison on an image from the Paris SteetView dataset~\cite{doersch2015makes}. Our Edge-LBAM is compared with (c) Global\&Local (GL)~\cite{IizukaGL}, (d) PConv~\cite{partialconv2017}, (e) DeepFillv2 (DFv2)~\cite{yu2018free}, (f) EdgeConnect (EC)~\cite{nazeri2019edgeconnect} and (g) MEDFE~\cite{liu2020rethinking}.}
	\label{fig:paris}
\end{figure*}

\begin{figure*}[hbt]
	\small
	\setlength{\tabcolsep}{2.0pt}
	\centering
	\begin{tabular}{cccccccc}
		\includegraphics[width=.130\textwidth]{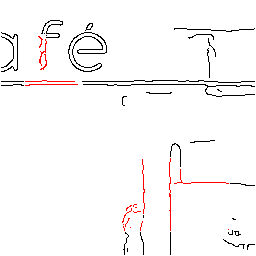}  &
		\includegraphics[width=.130\textwidth]{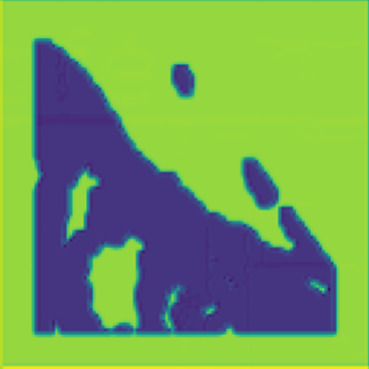}  &
		\includegraphics[width=.130\textwidth]{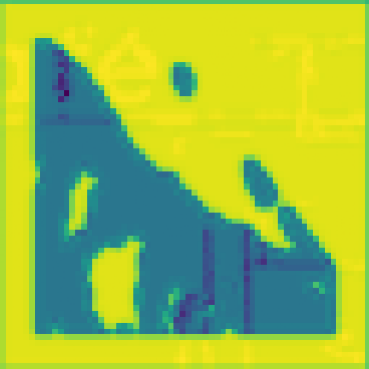}  &
		\includegraphics[width=.130\textwidth]{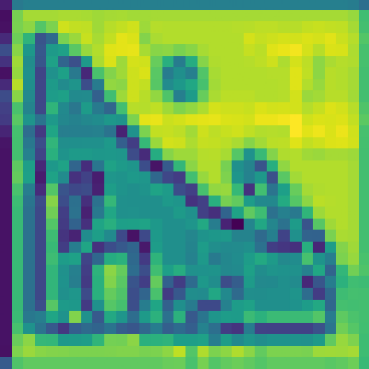}  &
		\includegraphics[width=.130\textwidth]{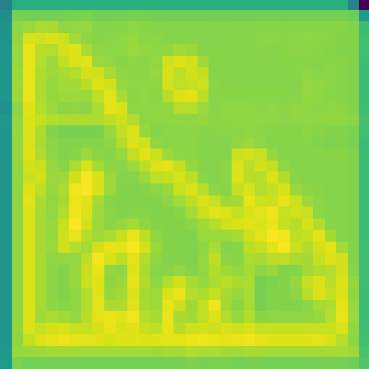}  &
		\includegraphics[width=.130\textwidth]{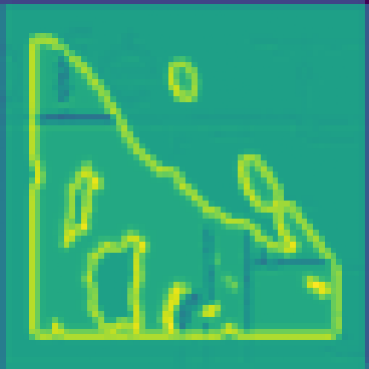}  &
		\includegraphics[width=.130\textwidth]{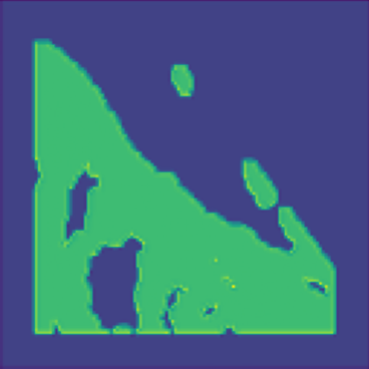} &
		\includegraphics[height=.130\textwidth]{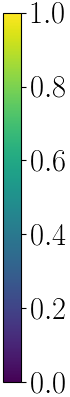}\\
	    \scriptsize{(a)} & \scriptsize{(b)} & \scriptsize{(c)} &\scriptsize{(d)} & \scriptsize{(e)} & \scriptsize{(f)} & \scriptsize{(g)} & \\
		\vspace{-2mm}\\	
	\end{tabular}
	
	\caption{Visualization of updated mask maps for different layers of our Edge-LBAM. (a) Estimated edge map for the input in Fig.~\ref{fig:paris}(b), (b)(c)(d) forward mask maps for the first three (\ie, 1, 2, and 3) layers, (e)(f)(g) reverse mask maps from the last three (\ie, 11, 12, and 13) layers.
}
		\label{fig:paris_visual}
	\end{figure*}

\IEEEraisesectionheading{\section{Introduction}\label{sec:introduction}}

Image inpainting~\cite{BertalmioInpainting} filling in missing pixels (\ie, holes) with plausible hypotheses is a challenging low level vision task with many real-world applications, \eg, distracting object removal, occluded region completion, and image editing.
Traditional image inpainting methods~\cite{criminisi2004region,Barnes:2009:PAR,XuPatchSparsity} usually adopt exemplar-based techniques which progressively fill in holes by searching and copying similar patches from known regions.
However, exemplar-based inpainting is limited in capturing high-level semantics and may fail in hallucinating complex and non-repetitive structures.
In contrast, deep convolutional networks (CNNs) are powerful in learning high-level semantics, and have greatly facilitated the advance of image inpainting in the recent few years~\cite{pathakCVPR16context,IizukaGL,Yang_2017_CVPR,song_contextual_2018, Yan_2018_Shift,yu2018generative}.

Existing CNN-based approaches~\cite{pathakCVPR16context,IizukaGL,Yang_2017_CVPR, song_contextual_2018,Yan_2018_Shift,yu2018generative} have achieved notable progress in the inpainting of rectangular regions.
However, most of them usually adopt standard convolution operation which treat valid pixels and holes indistinguishably, making them limited in handling irregular holes and leading to result with color discrepancy and blurriness (see Fig.~\ref{fig:paris}(c)).
The methods~\cite{song_contextual_2018,Yan_2018_Shift,yu2018generative} exploit the mask of holes to guide the feature propagation from known regions for enhancing decoder features in the holes, but are still insufficient in handling irregular holes.
Recently, partial convolution (PConv)~\cite{partialconv2017} has been introduced for coping with irregular holes while preventing color discrepancy and blurriness.
A PConv layer generally involves three steps, \ie, masked convolution, feature re-normalization, and mask-updating, which collaborate to make the convolution and re-normalization conditioned on only valid pixels and to gradually shrink the holes.
Nonetheless, the image structure information and inpainting confidence (\ie, filling-in confidence) are also not fully exploited.

In this paper, we present an edge-guided learnable bidirectional attention map (Edge-LBAM) to overcome the limitations of existing methods in handling irregular holes.
First, masked convolution, feature re-normalization and mask-updating in PConv~\cite{partialconv2017} are based on the hard 0-1 mask while completely trusting all filling-in intermediate features.
However, the filling-in confidence may vary with spatial position and convolution layers.
For example, in~\cite{song_contextual_2018} it is assumed that the filling-in features near hole boundary is more confident.
To address this issue, we revisit PConv without bias and show that masked convolution can be substituted with standard convolution.
And feature re-normalization can be interpreted as an attention mechanism where the attention map is generated based on the updated hard 0-1 mask.
Thus, we present a learnable forward attention map (LFAM) module for learning feature re-normalization and mask-updating.
Instead of hard 0-1 mask, LFAM can learn soft attention map for feature re-normalization.
Benefited from end-to-end training, LFAM is effective in adapting to irregular holes and learning mask-updating, thereby benefiting inpainting performance.

Moreover, filling-in order plays a pivotal role in exemplar-based inpainting~\cite{criminisi2004region,Barnes:2009:PAR,XuPatchSparsity} and largely depends on image structures.
Generally, better inpainting result can be obtained by filling in structures and other missing regions separately.
However, the filling-in order for PConv is purely determined by the input mask of holes,
while several other deep inpainting approaches first apply a rough inpainting stage, which are then used to guide the propagation of detailed information from known regions to
the holes~\cite{Yang_2017_CVPR,song_contextual_2018,Yan_2018_Shift,yu2018generative}.
Motivated by exemplar-based inpainting, we present edge-guided learnable forward attention map (Edge-LFAM) which utilizes both the mask of holes and edge map for mask-updating, thereby making the filling-in order to be structure-aware.
A multi-scale edge completion network is further suggested for predicting the coherent edge map from the input corrupted image.
Fig.~\ref{fig:paris_visual}{(b)$\sim$(d)} illustrates the visualization of updated masks from different encoder layers.
Interestingly, the mask of holes first shrinks in all directions until it meets the predicted edges.
Then, the model begins to fill in the missing structural regions, \ie, shrinking the holes only along the direction of predicted edges.
Albeit our method adopts a different filling-in order with exemplar-based inpainting, it can also effectively avoid filling in structural regions with background pixels.

%
%
%
%

%

Furthermore, PConv only considers the mask-updating in the encoder, while all-ones mask is simply adopted for decoder features.
Actually, the encoder features of known region will be concatenated with the decoder feature map for the U-Net structure~\cite{UNetRFB15a}.
Thus, it is reasonable to make the decoder concentrate more on the holes instead of known region.
To this end, we introduce edge-guided learnable reverse attention map (Edge-LRAM), which is then incorporated with Edge-LFAM to formulate our edge-guided learnable bidirectional attention maps (Edge-LBAM).
From Fig.~\ref{fig:paris_visual}, it can be seen that the masks in decoder emphasize more on the boundaries of the holes and gradually assign zeros to the known regions in the last layer.
We also empirically find that Edge-LBAM is beneficial to network training, making it feasible to include adversarial loss for improving visual quality of the inpainting result.


Extensive experiments are conducted on the Paris SteetView~\cite{doersch2015makes}, Places~\cite{zhou2017places} and CelebA-HQ256~\cite{karras2017progressive} datasets.
Qualitative and quantitative results show that our Edge-LBAM performs favorably against the state-of-the-art deep inpainting methods~\cite{IizukaGL,partialconv2017,yu2018free,nazeri2019edgeconnect,liu2020rethinking}.
As illustrated in Fig.~\ref{fig:paris}, our Edge-LBAM is more effective in preventing
color discrepancy and blurriness than Global\&Local (GL)~\cite{IizukaGL}, PConv~\cite{partialconv2017}, DeepFillv2~\cite{yu2018free} {and MEDFE~\cite{liu2020rethinking}}.
In comparison to the two-stage model EdgeConnect (EC)~\cite{nazeri2019edgeconnect}, our Edge-LBAM can utilize the predicted edges in mask-updating, and gets decent performance on the inpainting of structural regions while avoiding the interference of background pixels.

%
%

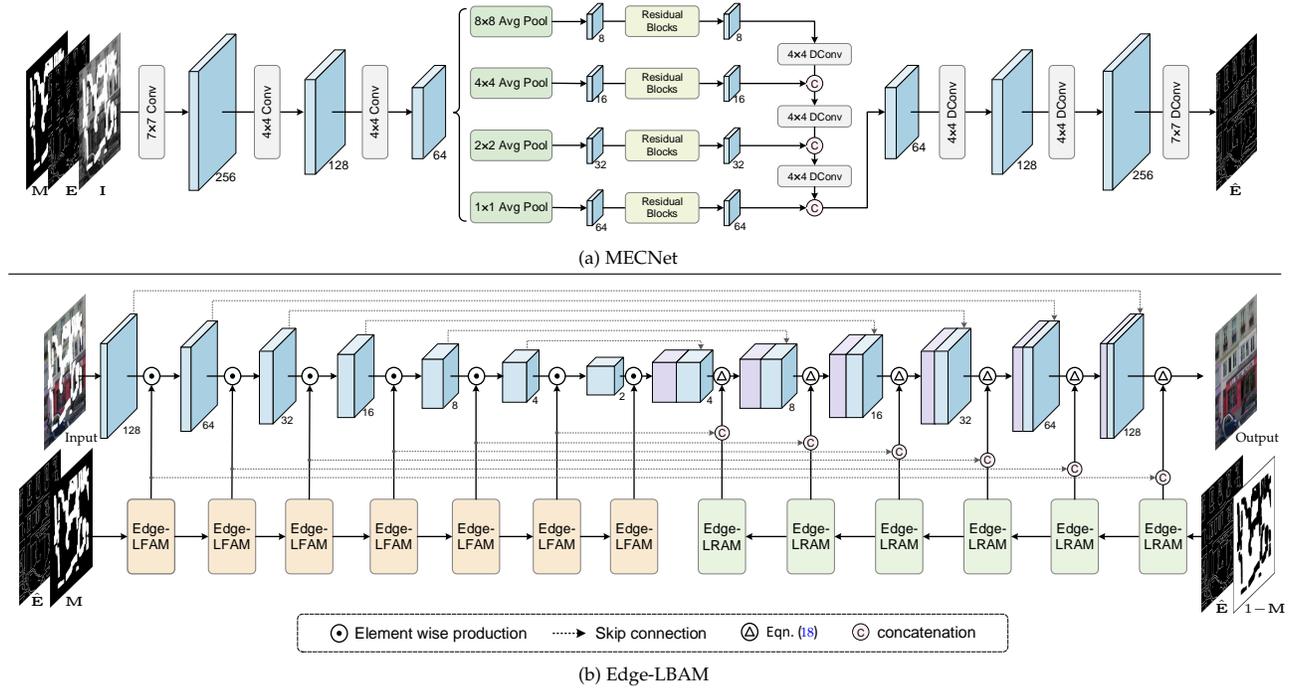
\begin{figure*}
	\begin{overpic}[width=1\textwidth]{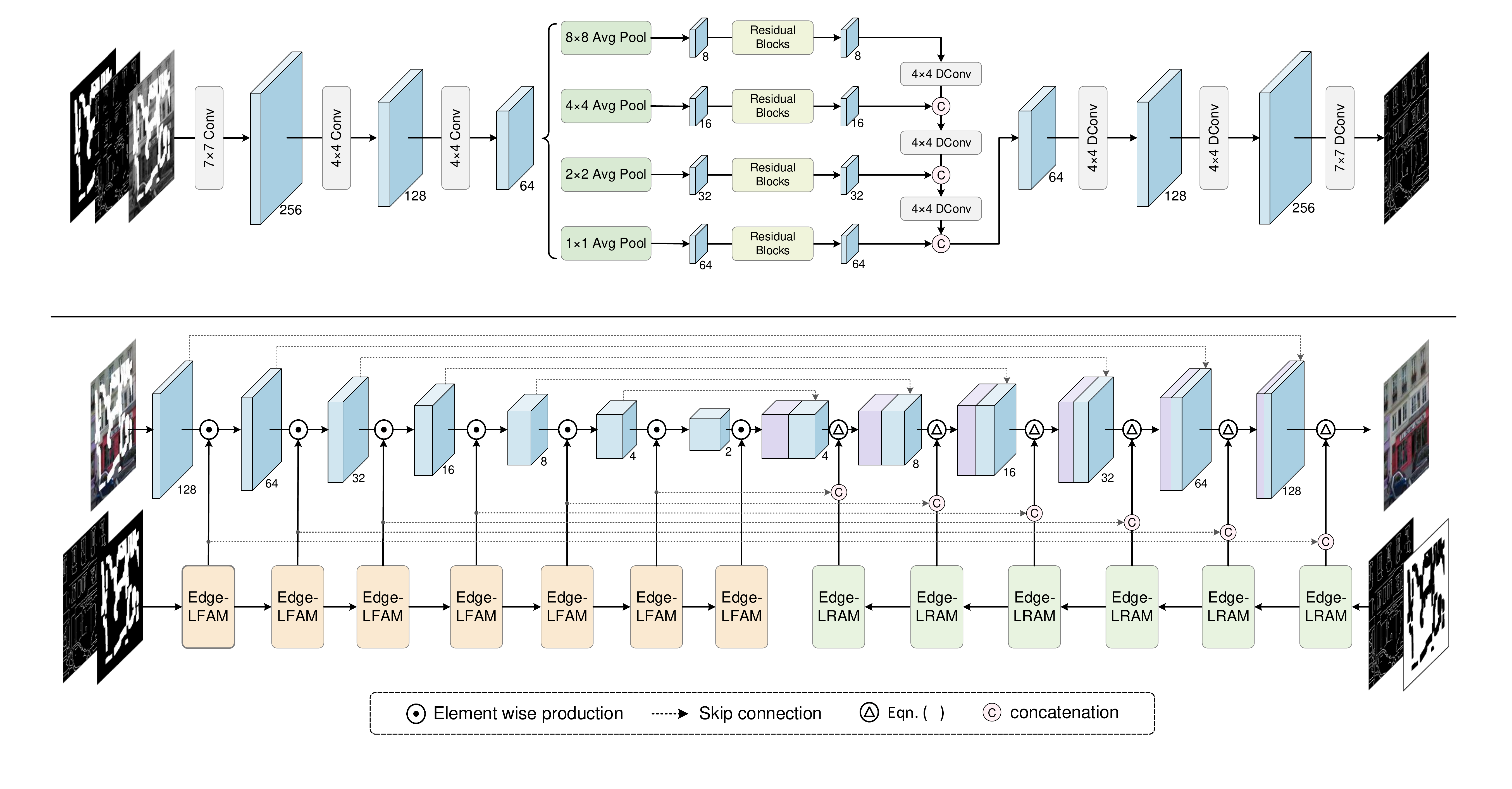}
		\put(60.95,4.7){\tiny ~\ref{renorm_lram}}
		\put(45,32){\scriptsize{(a) MECNet}}
		\put(10,37){\tiny $\mathbf{I}$}
		\put(5,37){\tiny $\mathbf{M}$}
		\put(7.5,37){\tiny $\mathbf{E}$}
		\put(92.5,37){\tiny $\hat{\mathbf{E}}$}
		\put(45,1.5){\scriptsize{(b) Edge-LBAM}}
		\put(7.5,19){\tiny Input}
		\put(93,19){\tiny Output}
		\put(5,7){\tiny  $\hat{\mathbf{E}}$}
		\put(7.5,7){\tiny$\mathbf{M}$}
		\put(93.6,6.5){\tiny $1\mathbf{-M}$}
		\put(91.5,6.5){\tiny $\hat{\mathbf{E}}$}
	\end{overpic}
	\caption{{The overall architecture of our proposed method: (a) multi-scale edge completion network (MECNet), (b) edge-guided learnable bidirectional attention maps (Edge-LBAM).}}
	\label{model}
\end{figure*}

This paper is an extension of our previous work~\cite{xie2019image}.
In comparison to~\cite{xie2019image}, we utilize both the mask of holes and predicted edge map for mask-updating, resulting in our Edge-LBAM method.
Moreover, we introduce a multi-scale edge completion network for effective prediction of coherent edges.
Experiments further validate the effectiveness of edge-guidance and multi-scale edge completion.
The contributions of this work are summarized as follows,
\begin{itemize}
	\item Instead of hard 0-1 mask and handcrafted mask-updating in
		  PConv~\cite{partialconv2017}, we present learnable forward attention map for better inpainting of irregular holes and propagating the mask along convolution layers.
	\item Taking the filling-in order into account, we suggest a multi-scale edge completion
	      network to predict reliable edge map incorporating with the mask for structure-aware mask-updating, resulting in edge-guided learnable forward attention map (Edge-LFAM).
	\item For making the decoder emphasize more on filling in holes, we further introduce
		  edge-guided learnable reverse attention map (Edge-LRAM), and incorporate it with Edge-LFAM to constitute our Edge-LBAM approach.
	\item Extensive experiments show that our Edge-LBAM is effective in preventing color
	      discrepancy and blurriness while generating plausible image structures, and performs favorably against the state-of-the-art deep inpainting methods.
\end{itemize}

The remainder of this paper is organized as follows.
Section~\ref{sec:related_work} briefly reviews the relevant works on image inpainting.
Section~\ref{sec:our proposed method} presents our Edge-LBAM method for improving image inpainting of irregular holes.
Section~\ref{sec:exp} reports the experimental evaluation on Edge-LBAM, and Section~\ref{sec:conclusion} ends this work with several concluding remarks.

\section{Related Work}\label{sec:related_work}
In this section, we present a brief survey on the relevant works, including both traditional and deep CNN-based image inpainting methods.
For traditional image inpainting, we focus on exemplar-based methods, and especially emphasize the strategies for designating filling-in order.
As for deep models, after briefly surveying CNN-based inpainting methods, we also introduce the progress in handling irregular holes and exploiting multi-stage schemes.
%
%
%
%
\subsection{Traditional Image Inpainting}\label{section2.1}
%
Diffusion-based and exemplar-based methods are two representative categories of traditional image inpainting approaches~\cite{guillemot2013image}.
%
%
Diffusion-based methods~\cite{BertalmioInpainting} propagate local information from known regions to missing pixels.
It usually can be formulated as a variational model with smoothness regularization.
For preserving image structures, several variants~\cite{ballester2001filling,chan2001local} have been suggested to prefer the propagation in the directions of local image structures.
However, diffusion-based inpainting is only suited for handling small holes, and is limited in completing detailed textures and complex structures of large regions.

%

Exemplar-based inpainting fills in holes in a progressive manner by exploiting patch-based self-similarity priors of images~\cite{criminisi2004region,Efros1999iccv}.
This category of methods are more effective in recovering detailed textures and handling large holes, but suffer from heavy computational burden and complex structural incoherence.
To relieve computational cost, PatchMatch~\cite{Barnes:2009:PAR} has been presented for efficient patch searching, and Barnes \etal~\cite{Barnes2010eccv} further generalized it to $k$-nearest neighbors ($k$NN) matching.
For better inpainting complex structures, several approaches have been suggested to incorporate exemplar-based inpainting with global coherence measure~\cite{wexler2004cvpr,he2012eccv} and global optimization algorithms~\cite{KomodakisPriority,pritch2009iccv}.
Another type of solution is to designate proper propagation process by filling in structures with higher priority.
And a number of patch priority measures have been proposed by combining confidence and
data terms.
Confidence term is generally defined as the ratio of known pixels in the input patch.
Criminisi \etal ~\cite{criminisi2004region} introduced a gradient-based data term for first inpainting linear structure.
While Xu and Sun~\cite{XuPatchSparsity} suggested a sparsity-based data term to indicate structural patches.
Le Meur \etal~\cite{LeMeur2011examplar} further presented a tensor-based data term, \ie, the eigenvalue discrepancy of Di Zenzo$'$s structure tensor~\cite{DiZenzo1986gradient}.
Albeit exemplar-based inpainting is effective in filling in repetitive textures, it is still not sufficient in capturing high-level semantics, thereby being limited in recovering complex and non-repetitive structures.

\subsection{Deep CNN-based Image Inpainting}\label{section2.2}
%
Deep CNN-based image inpainting aims at learning a direct mapping to predict the missing pixels.
While early CNN-based methods~\cite{XieDenoiseCNN,KohlerSSHJR2014,RenShepardConv} were presented for the inpainting of small and thin holes, most recent efforts have been devoted for more challenging issues, \ie, larger and irregular holes.
Benefited from its power representation ability and large scale end-to-end training, deep CNN generally is effective in recovering high-level semantic structures.
And several network architectures as well as losses have been presented.
In their seminal work, Pathak \etal~\cite{pathakCVPR16context} incorporated reconstruction and adversarial losses to train an encoder-decoder network.
Iizuka \etal~\cite{IizukaGL} deployed both global and local discriminators for respectively enforcing semantically plausible structures and photo-realistic details of inpainting result.
Wang \etal~\cite{WangMulticolumn} suggested to exploit confidence-driven reconstruction loss with implicit diversified MRF (ID-MRF) term to train a multi-column CNN.

To incorporate self-similarity image priors into CNN-based inpainting, Yang \etal~\cite{Yang_2017_CVPR} presented a two-stage method.
After structured content prediction, multi-scale neural patch synthesis (MNPS) is used for jointly optimizing the holistic content and local texture constraints in the mid-layer representation.
To relieve the computational cost of MNPS, contextual attention~\cite{yu2018generative} and patch-swap~\cite{song_contextual_2018} have been proposed for explicitly exploiting non-local self-similarity with deep feed-forward networks.
Yan \etal~\cite{Yan_2018_Shift} introduced a one-stage inpainting network, \ie, Shift-Net, where the decoder features in holes are use to guide the shift of encoder features from known regions to holes.
For pluralistic image completion, Zheng \etal~\cite{Zheng2019Pluralistic} suggested a network with two training paths, in which a reconstructive path is introduced to guide the learning of generative path during training.
And a short+long term attention layer is deployed to capture both intra-layer and inter-layer self-attention for improving inpainting result.
Liu \etal~\cite{liu2019coherent} presented a coherent semantic attention (CSA) layer by further considering the contextual similarity in feature space.
{In MEDFE~\cite{liu2020rethinking}, Liu \etal~adopted mutual encoder-decoder for recovering both semantic structures and texture details.
For feature equalization, bilateral propagation activation is suggested to extend non-local block by considering feature similarity and spatial distance.
%
}
However, for these attention-based models~\cite{yu2018generative,song_contextual_2018,Yan_2018_Shift,Zheng2019Pluralistic,liu2019coherent,liu2020rethinking}, the mask is only used to guide the feature propagation from known regions to holes, making them inadequate in handling irregular holes.

%

%
%


Liu \etal~\cite{partialconv2017} proposed a partial convolution (PConv) layer for effectively exploiting the mask of holes.
In PConv, the mask is used to both mask out the pixels in holes in convolution and to re-normalize the convolutional features.
While hard 0-1 mask-updating is adopted to gradually shrink the holes along the convolution layers.
Yu \etal~\cite{yu2018free} presented gated convolutions by generalizing PConv with learnable dynamic feature re-normalization.
Besides the corrupted image and mask, user sketch is also allowed to be fed into gated convolution for the end of free-form inpainting.
In comparison, our LBAM~\cite{xie2019image} concurrently generalized PConv with learnable attention for feature re-normalization and mask-updating.
And reverse attention map is further introduced to make the decoder concentrate more on filling in holes.

Multi-stage models have also been investigated for deep image inpainting.
To ease the difficulty of deep image inpainting, Zhang \etal~\cite{zhang2018mm} combined CNN with LSTM to form progressive generative network (PGN), and adopted a curriculum learning method.
Among existing multi-stage inpainting methods, one effective solution is to disentangle structure reconstruction and content hallucination.
Xiong \etal~\cite{Xiong_2019_CVPR} presented a three-stage foreground-aware inpainting method involving contour detection, contour completion and image completion.
Ren \etal~\cite{ren2019structureflow} proposed a two-stage model, where structure reconstruction is trained to predict the edge-preserved smooth images, and content hallucination is designed to yield image details based on the recovered smooth image.
Nazeri \etal~\cite{nazeri2019edgeconnect} also suggested a two-stage model EdgeConnect by first predicting salient edges and then generating inpainting result.
In contrast to EdgeConnect~\cite{nazeri2019edgeconnect}, our Edge-LBAM leverages the predicted edge map to produce structure-aware filling-in order, and thus is more effective in handling irregular holes and inpainting sematic structures.
Moreover, we also present a multi-scale edge completion network for better edge prediction.

\begin{figure*}
	\centering
	\begin{overpic}[width=1\textwidth]{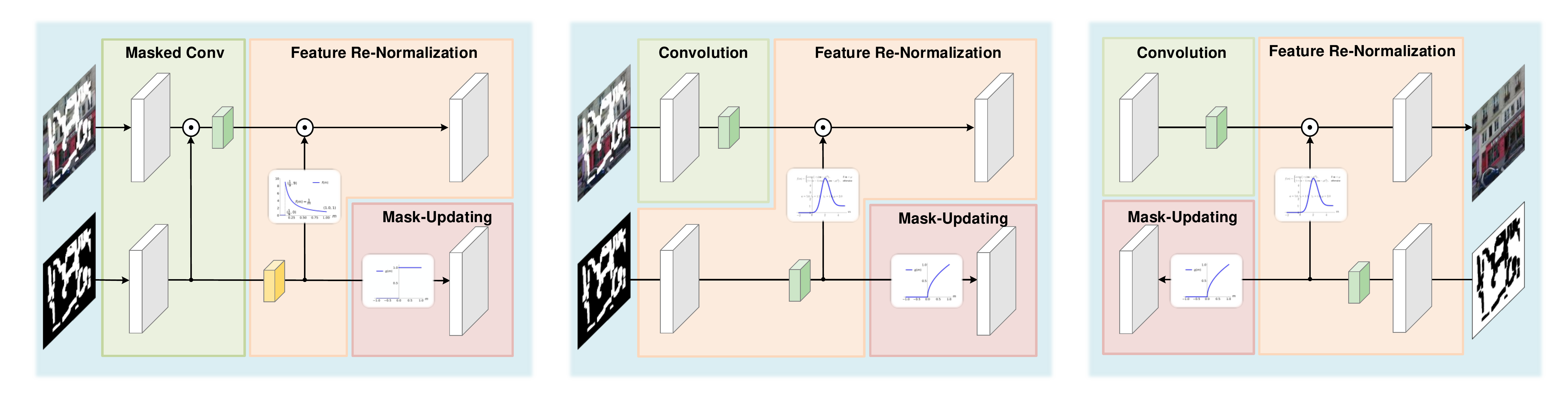}
		\put(15,0.5){\scriptsize{(a) PConv}}
		\put(3.5,13.4){\tiny{Input}}
		\put(4,4){\tiny{$\mathbf{M}$}}
		\put(9, 13.6){\tiny{$\mathbf{F}^{{in}}$}}
		\put(9, 4.2){\tiny{$\mathbf{M}^{in}$}}
		\put(13.5, 15.8){\tiny{$\mathbf{W}$}}
		\put(17.5, 11){\tiny{$f_{\scriptscriptstyle A}$}}
		\put(16.5, 6){\tiny{$\mathbf{k}_{{1}/{9}}$}}
		\put(24.3, 5.5){\tiny{$f_{M}$}}
		\put(29, 13.6){\tiny{$\mathbf{F}^{out}$}}
		\put(29, 4){\tiny{$\mathbf{M}^{out}$}}
		
		\put(48,0.5){\scriptsize{(b) LFAM}}
		\put(37.5,13.2){\tiny{Input}}
		\put(38,4){\tiny{$\mathbf{M}$}}
		\put(43, 14){\tiny{$\mathbf{F}^{in}$}}
		\put(43, 4.5){\tiny{$\mathbf{M}^{in}$}}
		\put(45.7, 15.8){\tiny{$\mathbf{W}$}}
		\put(50.5, 11){\tiny{$g_{A}$}}
		\put(50.5, 6){\tiny{$\mathbf{k}_{m}$}}
		\put(58.3, 5.5){\tiny{$g_M$}}
		\put(63, 13.6){\tiny{$\mathbf{F}^{out}$}}
		\put(62.6, 4){\tiny{$\mathbf{M}^{out}$}}
		
		\put(81,0.5){\scriptsize{(c) LRAM}}
		\put(95,13.4){\tiny Output}
		\put(76.5, 15.8){\tiny{$\mathbf{W_d}$}}
		\put(95,5){\tiny{$1\mathbf{-M}$}}
		\put(89.5,13.6){\tiny $\mathbf{F}^{out}_d$}
		\put(89.5,4){\tiny $\overline{\mathbf{M}}^{in}$}
		\put(72.2,13.8){\tiny $\mathbf{F}^{in}_d$}
		\put(72,4){\tiny $\overline{\mathbf{M}}^{out}$}
		\put(81.3,11){\tiny{ $g_A$}}
		\put(76,5.3){\tiny $g_M$}
		\put(85.4,6){\tiny {$\mathbf{k}_{\overline{m}}$}}

	\end{overpic}
	\caption{{Interplay models between mask and intermediate feature for PConv, LFAM and LRAM. Here, the white holes in $\mathbf{M}$ denotes missing region with value 0, and the black area denotes the known region with value 1.}}
	\label{LBAM}
\end{figure*}
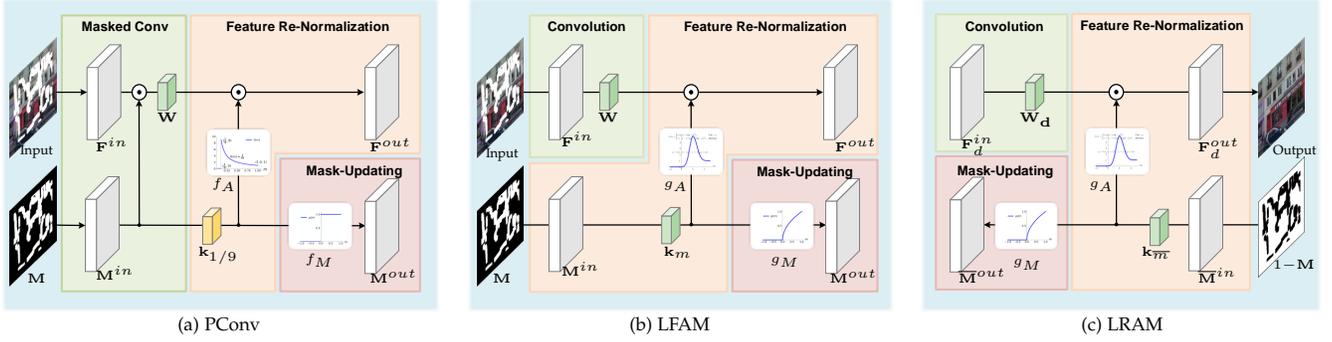

\section{Proposed Method}\label{sec:our proposed method}
In this section, we present the edge-guided learnable bidirectional attention maps (Edge-LBAM) for the effective image inpainting of irregular holes.
To begin with, we first revisit PConv~\cite{partialconv2017} and discuss its main limitations.
Then, we provide an overview of our Edge-LBAM and further describe its main components.


\subsection{Revisit Partial Convolution}\label{sec3.1}
In~\cite{partialconv2017}, Liu \etal~presented a partial convolution (PConv) method for handling irregular holes.
In general, a PConv layer takes the input feature map $\mathbf{F}^{in}$ and binary mask $\mathbf{M}^{in}$ as the input, and includes three steps, \ie, masked convolution, feature re-normalization, and mask-updating.
Denote by $\mathbf{W}$ the convolution filter and $b$ the bias.
To begin with, we define the convolved mask $\mathbf{M}^c$ as,
\begin{equation}\label{eqn:conv_mask}
	\mathbf{M}^c = \mathbf{M}^{in} \otimes \mathbf{k}_{{1}/{9}},
\end{equation}
where $\otimes$ denotes the convolution operator, $\mathbf{k}_{{1}/{9}}$ denotes a $3 \times 3$ convolution filter with each element {$({1}/{9})$}.
%
%
%
%
%
%
%
Then, masked convolution can be given by,
\begin{equation}\label{PConv-1}
	\mathbf{F}^{c} = \mathbf{W}^{T}(\mathbf{F}^{in} \odot \mathbf{M}^{in}),
\end{equation}
where $\mathbf{W}$ is the convolutional kernel for $\mathbf{F}^{in}$.
While feature re-normalization and mask-updating can be respectively formulated as,
\begin{equation}\label{PConv-2}
	\mathbf{F}^{out} =
	\begin{cases}
		\mathbf{F}^{c} \odot f_A(\mathbf{M}^c) + b, & \mbox{if}\ \mathbf{M}^c > 0 \\
		0, & \mbox{otherwise}
	\end{cases}
\end{equation}
\begin{equation}\label{PConv-3}
	\mathbf{M}^{out} = f_M(\mathbf{M}^c)
\end{equation}
Here, the activation functions $f_A(\mathbf{M}^c)$ and $f_M(\mathbf{M}^c)$ are respectively defined as,
\begin{equation}\label{Activation_A}
	f_A(\mathbf{M}^c) =
	\begin{cases}
		\frac{1}{\mathbf{M}^c}, & \mbox{if} \ \mathbf{M}^c > 0 \\
		0, & \mbox{otherwise}
	\end{cases}
\end{equation}
\begin{equation}\label{Activation_M}
	f_M(\mathbf{M}^c) =
	\begin{cases}
		1, & \mbox{if} \ \mathbf{M}^c > 0 \\
		0, & \mbox{otherwise}
	\end{cases}
\end{equation}

However, there remains several limitations with PConv in network training and handling irregular holes on images with complex semantic structures.
(i) The activation functions $f_A(\mathbf{M}^c)$ and $f_M(\mathbf{M}^c)$ are non-differential, thereby increasing the difficulty of end-to-end training.
The convolution filter {$\mathbf{k}_{{1}/{9}}$} results in handcrafted mask-updating, and thus cannot guarantee to be optimized for image inpainting.
(ii) From Eqns.~(\ref{eqn:conv_mask})$\&$(\ref{PConv-3}), the filling-in order for PConv is determined by the mask-updating rule and solely relies on the input mask of holes.
For exemplar-based inpainting, filling-in order is a crucial issue and global structures are generally required to be inpainted with higher priority.
Thus, it is worthy to improve PConv by making mask-updating to be structure-aware.
(iii) Last but not least, masked convolution and mask-updating are only considered for the encoder, and all-ones mask is simply adopted for decoder features.
Actually, encoder feature of known region will be concatenated with decoder feature.
Thus, proper masks are expected to be beneficial to make the decoder mainly concentrate on filling in the holes.

%

\subsection{Overview of Edge-LBAM}\label{sec3.2}
Without loss of generality, we define the attention map $\mathbf{A} = f_A(\mathbf{M}^c)$ and assume the bias $b = 0$~\cite{isola2017cvpr,Yan_2018_Shift}.
As illustrated in Fig.~\ref{LBAM}(a),
PConv can then be explained as a special Conv layer with pre- and post-convolution attention-based normalization.
Before feeding to the convolution, the input feature map $\mathbf{F}^{in}$ is first normalized based on the input mask $\mathbf{M}^{in}$, \ie, $\mathbf{F}^{in} \odot \mathbf{M}^{in}$.
After the convolution, $\mathbf{F}^c$ is further re-normalized based on the attention map $\mathbf{A}$.
From Eqn.~(\ref{Activation_A}), one can directly have $\mathbf{F}^{out} = \mathbf{F}^{c} \odot \mathbf{A}$.
According to Eqn.~(\ref{PConv-2}), the features in unfilled regions are guaranteed to be zeros.
Thus, we can further ignore pre-convolution normalization and only focus on post-convolution attention-based normalization.

From the above attention-based normalization perspective, we present an edge-guided learnable bidirectional attention map (Edge-LBAM) to overcome the limitations described in Sec.~\ref{sec3.1}.
For (i), we can design differential activation functions $f_A(\mathbf{M}^c)$ and $f_M(\mathbf{M}^c)$, and make the convolution filter $\mathbf{k}_{{1}/{9}}$ to be learnable, resulting in our learnable forward attention map (LFAM) module.
For (ii), we further extend LFAM to edge-guided learnable forward attention map (Edge-LFAM).
In particular, both the input mask and edge maps are fed into Edge-LFAM for making mask-updating to be structure-aware.
And a multi-scale edge completion network is developed for predicting reliable and coherent edge map.
As for (iii), we further provide edge-guided learnable reverse attention map (Edge-LRAM) for allowing the decoder emphasize more on the holes.
Finally, Edge-LRAM and Edge-LFAM are incorporated to form our Edge-LBAM method.

%
%
%
%
%

Fig.~\ref{model} illustrates the overall architecture of our proposed method, which generally involves two stages.
In the first stage, a multi-scale edge completion network is deployed to predict the edge map from the mask of holes, the corrupted image and edge map.
In the second stage, the Edge-LFAM and Edge-LRAM modules are collaborated with a U-Net backbone to constitute our Edge-LBAM model.
From Fig.~\ref{model}(b), the corrupted image is fed into the U-Net, while the predicted edge map and mask of holes are fed into the first Edge-LFAM module.
Besides, the last Edge-LRAM module also takes the predicted edge map and reverse mask as the input for the attention-based normalization of decoder feature.
Each of the Edge-LFAM and Edge-LRAM modules not only produces the attention map for feature re-normalization, but also generates the updated mask map and edge map for the next Edge-LFAM or preceding Edge-LRAM module.
%
%
%
%
The U-Net~\cite{UNetRFB15a} involves 14 layers by removing the bottleneck layer.
An Edge-LFAM module is incorporated with each of the first six encoder layers, while an Edge-LRAM module is applied to each of the last six decoder layers.
In the U-Net backbone, we use convolution kernel with the size of $4\times4$, stride 2 and padding 1, and set the bias to be zero.
After feature re-normalization, we apply batch normalization and leaky ReLU nonlinearity, and \emph{tanh} nonlinearity is employed right after convolution for the last layer.

%
%

In the following, we further introduce the main components of our proposed method, \ie, multi-scale edge completion, Edge-LFAM and Edge-LRAM, and describes the loss terms for training Edge-LBAM.

\begin{figure*}[htbp]
	\centering
	\begin{overpic}[width=1\textwidth]{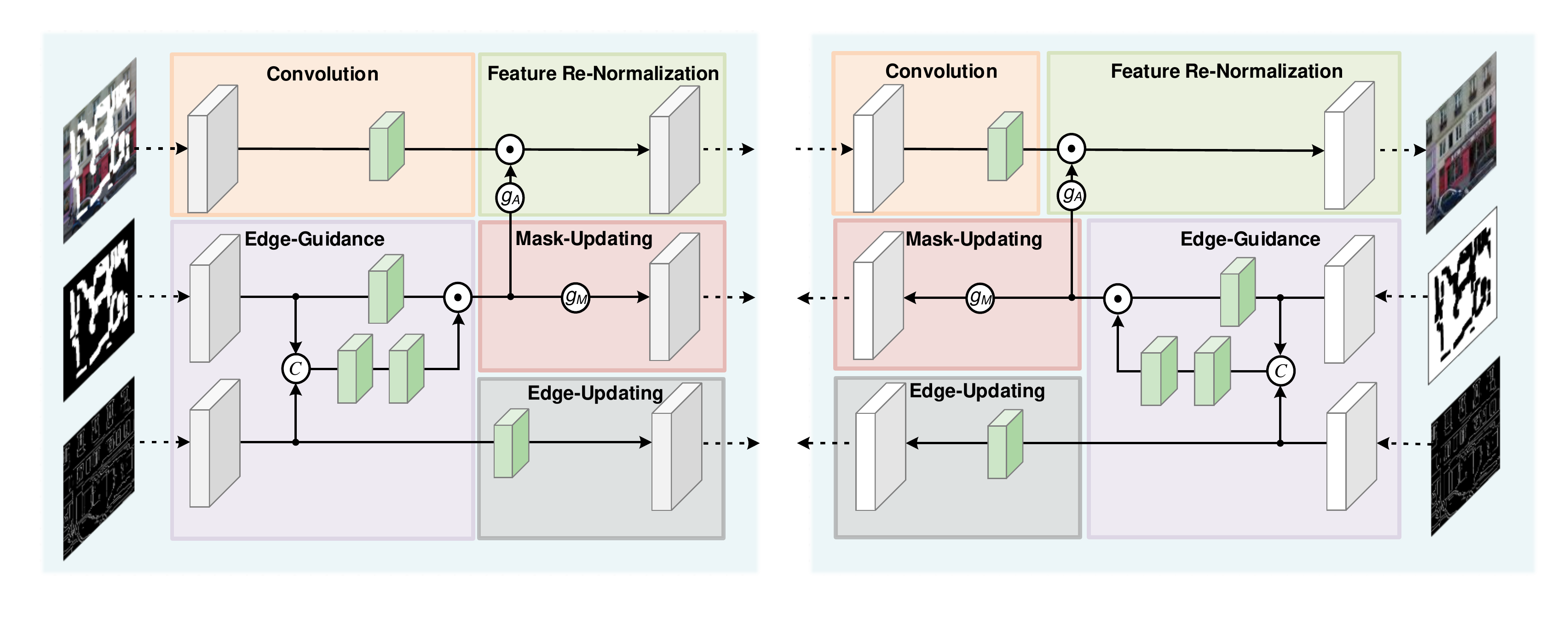}

		\put(20,1.5){\scriptsize{(a) Edge-LFAM}}
		\put(7,27){\tiny  Input}
		\put(7,17){\tiny $\mathbf{M}$}
		\put(7,6.5){\tiny $\hat{\mathbf{E}}$}
		\put(14,27){\tiny  $\mathbf{F}^{in}$}
		\put(14,17){\tiny  $\mathbf{M}^{in}$}
		\put(14,8){\tiny  $\mathbf{E}^{in}$}
		\put(25,28.5){\tiny  $\mathbf{W}$}
		\put(25,19.5){\tiny  $\mathbf{k}_{m}$}
		\put(23,14.5){\tiny  $\mathbf{k}_{1}$}
		\put(26.5,14.5){\tiny  $\mathbf{k}_{2}$}
		\put(33.2,10){\tiny  $\mathbf{k}_{e}$}
		\put(43.3,27){\tiny  $\mathbf{F}^{out}$}
		\put(42.8,17){\tiny  $\mathbf{M}^{out}$}
		\put(43,7.5){\tiny  $\mathbf{E}^{out}$}
		
		\put(70,1.5){\scriptsize{(b) Edge-LRAM}}
		\put(86.6,27.3){\tiny  $\mathbf{F}^{out}_d$}
		\put(86.5,17){\tiny $\overline{\mathbf{M}}^{in}$}
		\put(86.5,7){\tiny  $\mathbf{E}^{in}_d$}
		\put(56.5,27.3){\tiny  $\mathbf{F}^{in}_d$}
		\put(56.5,17.5){\tiny  $\overline{\mathbf{M}}^{out}$}
		\put(56.5,7.4){\tiny  $\mathbf{E}^{out}_d$}
		\put(64,28.5){\tiny  $\mathbf{W_d}$}
		\put(79.4,19.5){\tiny  $\mathbf{k}_{\overline m}$}
		\put(78,14.5){\tiny  $\mathbf{k}_{\overline 1}$}
		\put(74.4,14.5){\tiny  $\mathbf{k}_{\overline 2}$}
		\put(64.2,9.5){\tiny  $\mathbf{k}_{\overline e}$}
		\put(94,28){\tiny Output}
		\put(93.5,17.7){\tiny  {$1\mathbf{-M}$}}
		\put(94,7.5){\tiny $\hat{\mathbf{E}}$}
	\end{overpic}
	\caption{Illustration of our edge-guided learnable forward attention maps (Edge-LFAM) and reverse attention maps (Edge-LRAM), which contain five modules, \ie, Convolution, Feature Re-Normalization, Edge-Guidance, Mask-updating and Edge-Updating.}
	\label{edge-guide}
\end{figure*}
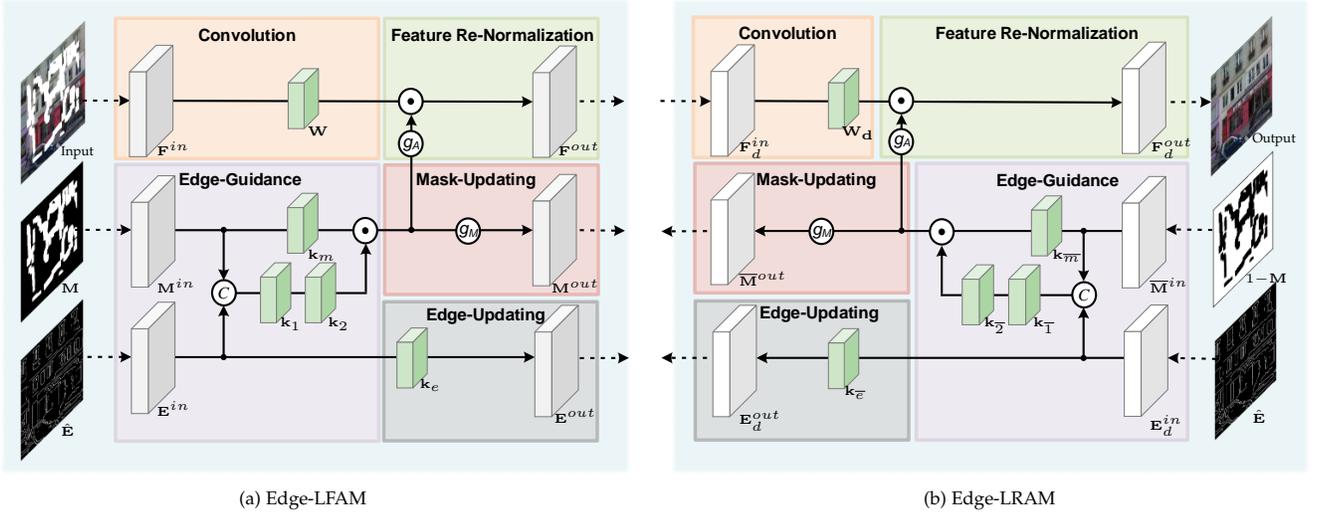

\subsection{Multi-scale Edge Completion}\label{sec2}
Motivated by exemplar-based inpainting, the filling-in order is generally assumed to be structure-aware.
To this end, we introduce a multi-scale edge completion network (MECNet) for predicting coherent edge map from the corrupted image.
As shown in Fig.~\ref{model}(a), the MECNet takes the mask of holes, the corrupted
image and incomplete edge map as the input to predict the edges in holes.
%
%
%
%
Given a corrupted image $\mathbf{I}$ with the mask of holes $\mathbf{M}$, we use the Canny edge detector on $\mathbf{I}$ to obtain the incomplete edge map $\mathbf{E}$.
To exploit the multi-scale representation, after two convolution layers, the features are pooled into different scales, \ie, 8, 16, 32 and 64, and respectively fed into four branches of eight residual blocks.
A deconvolution layer is deployed to upsample the feature from each lower-scale branch, which is then concatenated with the feature map from the next higher-scale branch.
Finally, another three deconvolution layers are applied to the highest scale to generate the edge completion result.


We incorporate the reconstruction loss and the adversarial loss to form the learning objective for training the MECNet $\hat{\mathbf{E}} = G_{MEC}(\mathbf{I}, \mathbf{M}, \mathbf{E})$.
Here, the network output $\hat{\mathbf{E}}$ is the estimated edge map.
Denote by $\mathbf{I}_{gt}$ the ground-truth image.
We use the Canny edge detector on $\mathbf{I}_{gt}$ to obtain the ground-truth edge map $\mathbf{E}_{gt}$.
%
%
%
%
To define the adversarial loss, we adopt the PatchGAN~\cite{isola2017cvpr,zhu2017unpaired} architecture as the discriminator $D^{MEC}$ for discriminating the predicted edge map from the real one in the patch level (\ie, 70$\times$70).
%
%
The adversarial loss is defined as,
\begin{equation}
	\begin{split}
		\mathcal{L}^{MEC}_{adv}  = \log\left(D^{MEC}(\mathbf{E}_{gt}, \mathbf{I})\right)  - \log\left(1 - D^{MEC}(\hat{\mathbf{E}}, \mathbf{I})\right).
	\end{split}
\end{equation}
%
%
Then, the reconstruction loss $\mathcal{L}_{rec}^{MEC}$ is defined as,
\begin{equation}\label{edge-l1}
	\begin{aligned}
		\mathcal{L}_{rec}^{MEC} =  \sum_{i=1}^{d} \parallel D_{l_i}^{MEC}(\hat{\mathbf{E}}, \mathbf{I}) -D^{MEC}_{l_i}(\mathbf{E}_{gt}, \mathbf{I}) \parallel_1, 
	\end{aligned}
\end{equation}
where $D_{l_i}^{MEC}$ denotes the $l_i$-th layer of the discriminator $D^{MEC}$.
We set $d = 5$ and $\{l_i\}_{i=1}^5 = \{1, 2, 3, 4, 5\}$. 
The learning objective for training MECNet can be given by,
\begin{equation}
	\mathcal{L}_{MEC} = \mathcal{L}^{MEC}_{adv} + \alpha_r \mathcal{L}_{rec}^{MEC},
\end{equation}
where $\alpha_r = 10$ is a regularization parameter for balancing the reconstruction and adversarial losses.

\subsection{Edge-Guided Learnable Forward Attention Maps}\label{sec3}
In this subsection, we first generalize PConv to present learnable forward attention map (LFAM).
Then, edge-guided learnable forward attention map (Edge-LFAM) is further introduced by incorporating both the mask of holes and edge map for mask-updating.

%

{\textbf{LFAM}.} As noted in Sec.~\ref{sec3.2}, masked convolution degenerates into standard convolution when the bias $b$ in Eqn.~(\ref{PConv-2}) equals zero.
Therefore, in our LFAM, only mask-updating and feature re-normalization are considered for generalizing PConv.
%
%
%
%
%
%
For mask-updating, we substitute the handcrafted convolution kernel $\mathbf{k}_{{1}/{9}}$ in Eqn.~(\ref{eqn:conv_mask}) with layer-wise and learnable convolution kernel $\mathbf{k}_{m}$, and the convolved mask in LFAM can then be obtained by,
\begin{equation}\label{eqn:conv_mask2}
	\mathbf{M}^c = \mathbf{M}^{in} \otimes \mathbf{k}_{m}.
\end{equation}
The convolution kernel $\mathbf{k}_{m}$ is learned from training data in an end-to-end manner, thereby allowing the mask in LFAM to be adaptive to irregular holes and propagation along with layers.

%

With $\mathbf{M}^c$, PConv adopts the activation function $f_M(\cdot)$ in Eqn.~(\ref{Activation_M}) for mask-updating.
However, $f_M(\cdot)$ is non-differential which hampers the gradient back-propagation during training.
Moreover, hard 0-1 masks are adopted in all layers of PConv, \ie, PConv indistinguishably trusts all filling-in intermediate features.
Nonetheless, filling-in features near hole boundary usually are assumed to be more confident~\cite{song_contextual_2018}.
Thus, our LFAM modifies the mask-updating rule by introducing a continuous activation function $g_M(\cdot)$,
\begin{equation}\label{Activation_gM}
	\mathbf{M}^{out} = g_M(\mathbf{M}^{c}) = \left(ReLU(\mathbf{M}^{c})\right)^{\alpha},
\end{equation}
where $\alpha \geq 0$ is a hyperparameter and we set $\alpha = 0.8$.
When $\mathbf{k}_{m} = \mathbf{k}_{{1}/{9}}$ and $\alpha = 0$, $g_M(\cdot)$ degenerates into $f_M(\cdot)$, and mask updating of our LFAM will be exactly the same as that of PConv.

Furthermore, instead of the non-differential $f_A(\cdot)$, we introduce an asymmetric Gaussian-shaped function as the activation function $g_A(\cdot)$ for feature re-normalization,
\begin{equation}\label{Activation_gA}
	g_A(\mathbf{M}^c) \!=\!
	\begin{cases}
		a \exp\left( - {\gamma_l (\mathbf{M}^c - \mu)^2} \right), & \mbox{if} \ \mathbf{M}^c \!<\! \mu \\
		1 \!+\! (a\!-\!1) \exp\left( - {\gamma_r (\mathbf{M}^c \!-\! \mu)^2} \right), & \mbox{else}
	\end{cases}
\end{equation}
where $a$, $\mu$, $\gamma_l$, and $\gamma_r$ are learnable parameters.
In our implementation, they are initialized as $a = 1.1, \mu=2.0,\gamma_l = 1.0, \gamma_r = 1.0$, and then updated during training in an end-to-end manner.
By introducing $\mathbf{A} = g_A(\mathbf{M}^c)$, the feature re-normalization in LFAM can then be accomplished by
\begin{equation}\label{renorm_lfam}
	\mathbf{F}^{out} = \mathbf{F}^{c} \odot \mathbf{A} = \mathbf{F}^{c} \odot g_A(\mathbf{M}^c).
\end{equation}
To sum up, Fig.~\ref{LBAM}(b) illustrates the three steps of LFAM, \ie, standard convolution, mask-updating, and feature re-normalization.
We also note that the differential $g_m(\cdot)$ and $g_A(\cdot)$ are crucial to LFAM.
For PConv, $\mathbf{k}_{{1}/{9}}$ is handcrafted and fixed during training, and thus the non-differential $f_m(\cdot)$ and $f_A(\cdot)$ can be acceptable.
But for LFAM, $g_m(\cdot)$ and $g_A(\cdot)$ are required to be differential for updating $\mathbf{k}_{m}$ via gradient back-propagation.

{\textbf{Edge-LFAM}.} For both PConv and LFAM, the filling-in order solely depends on the input mask of holes.
In exemplar-based inpainting, several patch priority measures~\cite{criminisi2004region,XuPatchSparsity,LeMeur2011examplar} are presented for filling in structures with higher priority, which are difficult to be deployed to deep image inpainting.
DeepFillv2~\cite{yu2018free} is the most relevant to our method, where user sketch can serve as structural guidance and is concatenated with the corrupted image and mask for learning dynamic feature re-normalization.
However, user sketch is required to be supplied by the user, and the input concatenation usually is not sufficient in exploiting user guidance.
In comparison, we suggest to incorporate the predicted edge map with the mask of holes for structure-aware mask-updating, resulting in our Edge-LFAM.

As shown in Fig.~\ref{edge-guide}(a), our Edge-LFAM adopts the same standard convolution and feature re-normalization as LFAM, and modifies the mask-updating step.
Denote by $\mathbf{F}^{in}$, $\mathbf{M}^{in}$, and $\mathbf{E}^{in}$ the input feature map, mask map and edge map, respectively.
On the one hand, one convolutional layer is deployed on $\mathbf{M}^{in}$ to obtain an intermediate mask $\mathbf{M}^{int}$.
On the other hand, $\mathbf{M}^{in}$ and $\mathbf{E}^{in}$ are concatenated and passed through two convolutional layers to attain the edge attentional map $\mathbf{A}^{E}$.
The convolved mask $\mathbf{M}^{c}$ in Edge-LFAM is then defined as,
\begin{equation}\label{eqn::mc_lefam}
\mathbf{M}^{c} = \mathbf{M}^{int} \odot \mathbf{A}^{E}.
\end{equation}
In addition, the input edge map $\mathbf{E}^{in}$ is fed to a convolutional layer to get the output edge map $\mathbf{E}^{out}$.
Following LFAM, the output mask map $\mathbf{M}^{out}$ can be obtained by $\mathbf{M}^{out} = g_M(\mathbf{M}^{c})$, and the attention map $\mathbf{A} = f_A(\mathbf{M}^c)$ can then be exploited for feature re-normalization via Eqn.~(\ref{renorm_lfam}).

%

For Eqn.~(\ref{eqn::mc_lefam}), it can be seen that the mask-updating in Edge-LFAM depends on both input mask map $\mathbf{M}^{in}$ and edge map $\mathbf{E}^{in}$.
In particular, the intermediate mask map $\mathbf{M}^{int}$ is completely determined by $\mathbf{M}^{in}$.
Intuitively, $\mathbf{M}^{in}$ can be acceptable when the mask propagation is non-overlapped or codirectional with the predicted edges.
However, it should be prevented when the mask propagation is perpendicular to the predicted edges.
To this end, we employ two convolutional layers on the concatenation of $\mathbf{M}^{in}$ and $\mathbf{E}^{in}$ to generate $\mathbf{A}^{E}$ for indicating the regions of cross-edge mask propagation.
Consequently, the mask-updating in Edge-LFAM is structure-aware and is beneficial to inpainting result.

\subsection{Edge-Guided Learnable Reverse Attention Maps}\label{sec5}
The encoder-decoder structure is usually adopted in deep image inpainting~\cite{pathakCVPR16context, song_contextual_2018,Yan_2018_Shift}.
While LFAM and Edge-LFAM can be incorporated with the encoder for mask-updating and feature re-normalization.
As for PConv, all-ones mask is simply adopted for decoder features.
Considering that the encoder feature of known region will be concatenated with the feature map of decoder, we actually require the decoder to only concentrate more on the complementary regions.
To this end, this subsection first introduces learnable reverse attention map (LRAM), and then further presents edge-guided learnable reverse attention map (Edge-LRAM) for structure-aware mask-updating.

{\textbf{LRAM}.} To begin with, we introduce the complementary mask of $\mathbf{M}$ as $\overline{\mathbf{M}} = 1 - \mathbf{M}$, which is then used to generate the input mask $\overline{\mathbf{M}}^{in}$ for each LRAM module.
Analogous to LFAM, our LRAM also involves three steps, \ie, standard convolution, mask-updating, and feature re-normalization.
Using the $(L-l)$-th layer of the U-Net as an example, the input feature map to LRAM involves two parts, that from the encoder ${\mathbf{F}}^{in}_e$ and that from the decoder ${\mathbf{F}}^{in}_d$.
The standard convolution can then be given by,
\begin{equation}\label{LRAM-1}
	\mathbf{F}^{c}_e = \mathbf{W}_e^{T}\mathbf{F}^{in}_e, \,\, \mathbf{F}^{c}_d = \mathbf{W}_d^{T}\mathbf{F}^{in}_d,
\end{equation}
where $\mathbf{W}_e$ and $\mathbf{W}_d$ are the convolutional kernels for $\mathbf{F}^{in}_e$ and $\mathbf{F}^{in}_d$, respectively.
For mask-updating, the convolved mask $\overline{\mathbf{M}}^{c}$ for LRAM is defined as,
\begin{equation}\label{eqn:conv_mask3}
	\overline{\mathbf{M}}^{c} = \overline{\mathbf{M}}^{in} \otimes \mathbf{k}_{\overline{m}},
\end{equation}
and the updated mask map is then given by,
\begin{equation}\label{Activation_gM2}
	\overline{\mathbf{M}}^{out} = g_M(\overline{\mathbf{M}}^{c}).
\end{equation}
In terms of feature re-normalization, we use $\mathbf{A}_d = g_A(\overline{\mathbf{M}}^c)$ to generate the attention map for $\mathbf{F}^{c}_d$.
As for $\mathbf{F}^{c}_e$, we adopt the attention map $\mathbf{A}_e$ used in the corresponding LFAM module for computing ${\mathbf{F}}^{in}_e$.
Consequently, the feature re-normalization in LRAM is provided by,
\begin{equation}\label{renorm_lram}
	\mathbf{F}^{out}_d = \mathbf{F}^{c}_e \odot \mathbf{A}_e + \mathbf{F}^{c}_d \odot \mathbf{A}_d.
\end{equation}

Fig.~\ref{LBAM}(c) illustrates the three steps of LRAM.
Unlike LFAM, while $\mathbf{F}^{out}_d$ is fed into the next layer, \ie, the $(L-l+1)$-th layer, of the U-Net, $\overline{\mathbf{M}}^{out}$ will be taken as the input mask to the preceding LRAM module, \ie, the $(L-l-1)$-th layer.
This explain why it is named as learnable reverse attention map (LRAM).
Moreover, LRAM makes the decoder concentrate more on generating complementary features to encoder feature map, and thus is beneficial to network training and inpainting performance.

{\textbf{Edge-LRAM}.} We further present Edge-LRAM by incorporating the predicted edge map for improving the mask-updating of LRAM.
Fig.~\ref{edge-guide}(b) shows the main steps of a Edge-LRAM module, where the standard convolution and feature re-normalization are the same as those in LRAM.
Denote by $\overline{\mathbf{M}}^{in}$ and $\mathbf{E}^{in}_d$ the input mask map and edge map for Edge-LRAM, respectively.
Analogous to Edge-LFAM, we apply one convolutional layer on $\overline{\mathbf{M}}^{in}$, and obtain the intermediate mask map $\overline{\mathbf{M}}^{int}$.
Two convolutional layers are deployed on the concatenation of $\overline{\mathbf{M}}^{in}$ and $\mathbf{E}^{in}$ to get the edge attentional map $\mathbf{A}^{E}_d$.
The convolved mask $\overline{\mathbf{M}}^{c}$ in Edge-LRAM is defined as,
\begin{equation}\label{eqn::mc_leram}
\overline{\mathbf{M}}^{c} = \overline{\mathbf{M}}^{int} \odot \mathbf{A}^{E}_d.
\end{equation}
And the output mask map $\overline{\mathbf{M}}^{out}$ can be obtained by,
\begin{equation}\label{eqn::mupdate_leram}
\overline{\mathbf{M}}^{out} = g_M(\overline{\mathbf{M}}^{c}).
\end{equation}
Furthermore, the attention map $\mathbf{A}_d = f_A(\overline{\mathbf{M}}^c)$ is used for feature re-normalization.
In addition, the input edge map $\mathbf{E}^{in}_d$ is fed to a convolutional layer to get the output edge map $\mathbf{E}^{out}_d$.
For Edge-LRAM, it is noteworthy that both $\mathbf{E}^{out}_d$ and $\overline{\mathbf{M}}^{out}$ will be fed into the preceding Edge-LRAM module.

Actually, Edge-LBAM only requires to generate complementary features to the encoder feature map for improved inpainting result.
As shown in Fig.~\ref{fig:paris_visual}, for the last layer, the attention map $\mathbf{A}$ in Edge-LBAM focus only filling in the holes.
As for the inner layers, the attention map $\mathbf{A}$ begins to emphasize the boundaries of holes, in which encoder and decoder features are required to collaborate to attain better inpainting result.

%
%
%
%

\begin{table*}[!tbp]
	\scriptsize
	\begin{center}
		\caption{Quantitative comparision results on the Paris Street View~\cite{doersch2015makes}, Places~\cite{zhou2017places} and Celeba-HQ256~\cite{karras2017progressive} datasets with Global-Local (GL)~\cite{IizukaGL}, Partial Convolution (PConv)~\cite{partialconv2017}, DeepFillv2 (DFv2)~\cite{yu2018free}, EdgeConnect (EC)~\cite{nazeri2019edgeconnect} and MEDFE~\cite{liu2020rethinking}. For metric with $\downarrow$, lower result means better performance, whereas metric with $\uparrow$ means higher result is better.
		}
		\begin{tabular}{C{0.3cm} C{1.0cm} C{0.9cm} C{0.9cm} C{0.9cm} C{0.9cm} C{0.9cm} C{0.9cm} C{0.9cm} C{0.9cm} C{0.9cm} C{0.9cm} C{0.9cm} C{0.9cm}}
			\toprule
			~& ~\multirow{2}{*}{ Method} &\multicolumn{4}{c}{Paris Street View}&\multicolumn{4}{c}{Places} &\multicolumn{4}{c}{CelebA-HQ256}\\
			\cmidrule{3-14}
			~&~& 10-20\% & 20-30\% & 30-40\% &40-50\%&10-20\% & 20-30\% & 30-40\% &40-50\%  &10-20\% & 20-30\% & 30-40\% &40-50\% \\
			\midrule
			\midrule
			\multirow{5}*{\rotatebox{90}{PSNR$\uparrow$}} & GL &30.46 & 25.64 & 23.06 & 20.84 &30.12 & 25.03 & 22.92 & 20.65 & 31.07 & 26.17 & 24.64 & 21.64\\
			~&PConv&30.56 &26.38 & 24.10 &21.14 & 28.32 & 25.52 & 22.89 & 21.38 &31.76 & 27.10&25.03 &22.45\\
			~& DFv2 & 31.07 & 27.04 & 25.02 &23.17 & 30.04 & 26.21 & 24.35 & 22.31 & 32.20 & 27.71 & 26.16 & 23.83\\
			~& EC & 31.10 & 27.53 & 25.56 & 23.93 & 30.06 & 26.70 & 24.95 & 23.18 & 33.36 & 28.54 & 27.04 & 24.57\\
			~& MEDFE & 31.24 & 27.14 & 25.05 & 23.21 & 30.20 & 26.42 & 24.74 & 23.11 & - & - & - & -\\
			~&Ours& \textbf{32.17} & \textbf{27.96} & \textbf{25.86} & \textbf{24.28} & \textbf{31.09} & \textbf{27.52} & \textbf{25.68} & \textbf{23.74} & \textbf{33.47} & \textbf{28.66} & \textbf{27.06} &\textbf{24.62}\\
			\midrule
			\midrule
			\multirow{5}*{\rotatebox{90}{SSIM$\uparrow$}} & GL& 0.900 & 0.815 & 0.753 & 0.655 & 0.887 & 0.790 & 0.722 & 0.635 & 0.894 & 0.809 & 0.756 &0.663\\
			~&PConv& 0.902 &0.821 &0.758 &0.669 &0.876 & 0.763 & 0.657 & 0.572 &0.902 & 0.811 &0.759 &0.667\\
			~&DFv2&0.914 &\textbf{0.828} & 0.763 & 0.671 & 0.893 &0.798 &0.736 & 0.640 & 0.904 &0.813 &0.763 &0.676\\
			~&EC&0.906 &0.817 & 0.760 & 0.676 & 0.891 & 0.799& 0.732&0.652 &0.917 & 0.834&0.788& 0.697\\
			~&MEDFE&0.908 &0.812 & 0.750 & 0.661 & 0.889 & 0.796& 0.717&0.637 &- & -&-& -\\
			~&Ours&\textbf{0.915} & 0.822 & \textbf{0.764} & \textbf{0.677} &\textbf{ 0.895} & \textbf{0.802}&\textbf{0.737}&\textbf{0.656}&\textbf{0.921}&\textbf{0.835} &\textbf{0.780} &\textbf{0.703}\\
			\midrule
			\midrule
			\multirow{5}*{\rotatebox{90}{$l_1(\%)$$\downarrow$}}&GL&1.21&2.63&3.77&5.11& 1.24 & 2.80 & 3.79 & 5.31 & 1.01 & 2.46 & 3.05 & 4.66\\
			~&PConv&1.17 &2.29 &3.24 & 4.12& 1.43 & 2.38 & 3.09 & 5.22 &0.96 & 2.04 &2.75 & 3.55\\
			~&DFv2& 0.73 & 1.47 &2.03 &2.83 & 0.79 & 1.56 & 2.10 & 2.97&0.66&1.39 &1.81 & 2.65\\
			~&EC& 0.65 & 1.37 &1.92 &2.68 & 0.75 & \textbf{1.48}&2.08&2.92&\textbf{0.49}&1.15 & 1.59&2.38\\
			~&MEDFE& 0.68 & 1.42 &1.99 &2.87 & 0.83 & 1.66 &2.20&2.76&-&-&-&-\\
			~&Ours& \textbf{0.60} & \textbf{1.33} & \textbf{1.88} &\textbf{2.59} &\textbf{0.74} & 1.53&\textbf{2.03}&\textbf{2.91}&0.50&\textbf{1.13} &\textbf{1.57}&\textbf{2.35}\\
			\midrule
			\midrule
			\multirow{5}*{\rotatebox{90}{LPIPS$\downarrow$}} & GL&0.1026 &0.2082&0.3151&0.4397&0.1238 &0.2264&0.3261&0.4525&0.0569&0.1656&0.2321&0.3778\\
			~&PConv&0.0609&0.1329&0.2020&0.3021&0.0953&0.1826&0.2450&0.3673&0.0477&0.1534&0.2091&0.3313\\
			~&DFv2&0.0589 & 0.1280&0.1803& 0.2604 &0.0801&0.1512&0.2169&0.3067&0.0423&0.1295&0.1541&0.2657\\
			~&EC&0.0558 &0.1139&0.1592&0.2292&0.0772  &\textbf{0.1494}&0.2093&0.2984&0.0407&0.0889&\textbf{0.1168}&0.1813\\
			~&MEDFE&0.0603 &0.1342&0.1894&0.2740&0.0795 &0.1626&0.2310&0.3279&-&-&-&-\\
			~&Ours&\textbf{0.0235} &\textbf{0.0647}&\textbf{0.0941}&\textbf{0.1490}&\textbf{0.0713} &0.1537&\textbf{0.2079}&\textbf{0.2930}&\textbf{0.0386}&\textbf{0.0869}&0.1275&\textbf{0.1785}\\
			\bottomrule
		\end{tabular}
		\label{quantitive_results}
	\end{center}
\end{table*}	

\subsection{Loss Functions}\label{sec7}
For training Edge-LBAM, we incorporate pixel reconstruction loss $\mathcal{L}_{\ell_{1}}$, perceptual loss $\mathcal{L}_{perc}$~\cite{Johnson2016Perceptual}, style loss $\mathcal{L}_{style}$~\cite{Gatys2016ImageST} and adversarial loss $\mathcal{L}_{adv}$~\cite{Goodfellow_GAN} to constitute our learning objective,
\begin{equation}\label{joint-loss}
\mathcal{L} = \lambda_{1} \mathcal{L}_{\ell_{1}} + \lambda_{2} \mathcal{L}_{adv} + \lambda_{3} \mathcal{L}_{perc} + \lambda_{4} \mathcal{L}_{style}
\end{equation}
where $\lambda_{1}$, $\lambda_{2}$, $\lambda_{3}$, and $\lambda_{4}$ are the tradeoff parameters.
In our implementation, we empirically set $\lambda_{1} = 1$, $\lambda_{2} = 0.1$, $\lambda_{3} = 0.05$ and $\lambda_{4} = 120$.
In the following, we will further explain these loss terms.
%
%


{\textbf{Pixel Reconstruction Loss}.} Denote by $\mathbf{I}$ the input corrupted image, $\mathbf{M}$ the binary mask region, $\mathbf{I}^{gt}$ the ground-truth image, and $\hat{\mathbf{E}}$ the predicted edge map.
Let $\boldsymbol{\Theta}$ be the model parameters of Edge-LBAM.
Naturally, the output of Edge-LBAM $\hat{\mathbf{I}} = \Phi(\mathbf{I}, \mathbf{M}, \hat{\mathbf{E}}; {\boldsymbol\Theta})$ is required to approximate the ground-truth image $\mathbf{I}^{gt}$, which can be depicted as the $\ell_{1}$-norm pixel reconstruction loss,
\begin{equation}
\label{l1loss}
\mathcal{L}_{\ell_{1}} = {\parallel \hat{\mathbf{I}} - \mathbf{I}^{gt} \parallel}_{1}.
\end{equation}

%



{\textbf{Perceptual Loss}.} The perceptual loss is further introduced to capture the high-level perceptual similarity between inpainting result $\hat{\mathbf{I}}$ and ground-truth image $\mathbf{I}^{gt}$.
%
%
In particular, we define the perceptual loss $\mathcal{L}_{perc}$ on the VGG-16 network~\cite{SimonyanZ14a} pre-trained on ImageNet~\cite{ILSVRC15},
\begin{equation}\label{perc-loss}
\mathcal{L}_{perc} = \frac{1}{N}\sum\nolimits_{i=1}^{N} \parallel \mathcal{P}^{i}(\mathbf{I}^{gt}) - \mathcal{P}^{i}(\hat{\mathbf{I}})  \parallel^2,
\end{equation}
where $\mathcal{P}^{i}(\cdot)$ denotes the feature maps of the $i$-th pooling layer.
In our implementation, we adopt the $pool$-1, $pool$-2, and $pool$-3 layers of the pre-trained VGG-16.

%
{\textbf{Style Loss}.}  The style loss is first introduced in~\cite{Gatys2016ImageST} for neural style transfer, and has then been adopted in image super-resolution~\cite{Sajjadi_2017_ICCV} and image inpainting~\cite{partialconv2017} for better recovery of detailed textures.
Following~\cite{partialconv2017}, we define the style loss also on the $N$ pooling layers of VGG-16,
\begin{equation}\label{style-loss}
\begin{split}
\mathcal{L}_{style} = & \frac{1}{N} \sum\nolimits_{i=1}^{N} \frac{1}{C_{i}^2}  \times  \\
&\parallel \mathcal{P}^{i}(\mathbf{I}^{gt}) (\mathcal{P}^{i}(\mathbf{I}^{gt}))^{T} - \mathcal{P}^{i}(\hat{\mathbf{I}}) (\mathcal{P}^{i}(\hat{\mathbf{I}}))^{T} \parallel^2
\end{split}
\end{equation}
where $C_{i}$ denotes the number of channels for the $i$-th pooling layer, and $\mathcal{P}^{i}(\cdot) (\mathcal{P}^{i}(\cdot))^{T}$ indicates the Gram matrix for the $i$-th layer of feature map.


{\textbf{Adversarial Loss}.} The adversarial loss~\cite{Goodfellow_GAN} has been widely adopted in image generation~\cite{salimans2016improved,pixelCNN,han2017stackgan} and low level vision~\cite{Ledig2016a}, and is also adopted in Edge-LRAM for improving the visual quality of inpainting results.
We exploit the Wasserstein distance~\cite{MartinWGAN} for measuring distribution discrepancy, and formulate the adversarial loss as,
\begin{equation}
\label{adv-loss}
\mathcal{L}_{adv} =  \mathbb{E}_{\mathbf{I}^{gt} } \left[ D(\mathbf{I}^{gt}) \right]
 - \mathbb{E}_{\hat{\mathbf{I}} } \left[ D(\hat{\mathbf{I}}) \right],
\end{equation}
where $D(\cdot)$ denotes the discriminator.
Minimizing the adversarial loss encourages the inpainting result being indistinguishable from ground-truth image, and thus is beneficial to visual quality.
In our implementation, we use a two-column network with seven layers as the discriminator $D(\cdot)$.
In particular, one column takes the known regions as the input while another column takes the results of the holes as the input.
After six layers, the feature maps from the two columns are concatenated and fed into the 7-th layer.
Then, the discriminator $D(\cdot)$ can be updated in an adversarial manner by maximizing the adversarial loss, and gradient penalty~\cite{ishaan2017improved} is further introduced for enforcing the Lipschitz constraint,
\begin{equation}\label{adv-loss2}
\begin{split}
\max_{D}\,\, & \mathbb{E}_{\mathbf{I}^{gt} } \left[ D(\mathbf{I}^{gt}) \right]
 - \mathbb{E}_{\hat{\mathbf{I}} } \left[ D(\hat{\mathbf{I}}) \right]\\
& + \lambda \mathbb{E}_{\tilde{\mathbf{I}} }\left[(\parallel \nabla_{\tilde{\mathbf{I}}}D(\tilde{\mathbf{I}}) \parallel^2 - 1)^2 \right]
\end{split}
\end{equation}
where $\tilde{\mathbf{I}}$ is sampled from the linear interpolation of $\mathbf{I}^{gt}$ and $\hat{\mathbf{I}}$ with a randomly selected factor, and $\lambda$ is set to 10 in our experiments.

\section{Experiments}\label{sec:exp}
In this section, experiments are conducted to evaluate our Edge-LBAM qualitatively and quantitatively.
In Sec.~\ref{trainingdetails}, we introduce the training and test datasets, competing methods and implementation details.
Then, Sec.~\ref{compare} compares our Edge-LBAM with several state-of-the-art image inpainting methods, and ablation studies are carried out in Sec.~\ref{ablationstudy} to assess each component of our Edge-LBAM.
%
%
%
	
\begin{figure*}[hbt]
	\small
	\setlength{\tabcolsep}{2.0pt}
	\centering
	\begin{tabular}{ccccccc}
		
		\includegraphics[width=.135\textwidth]{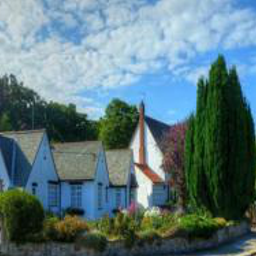}  &
		\includegraphics[width=.135\textwidth]{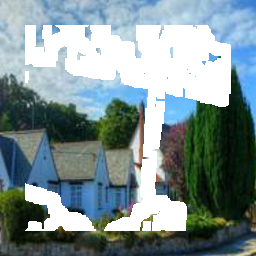}  &
		\includegraphics[width=.135\textwidth]{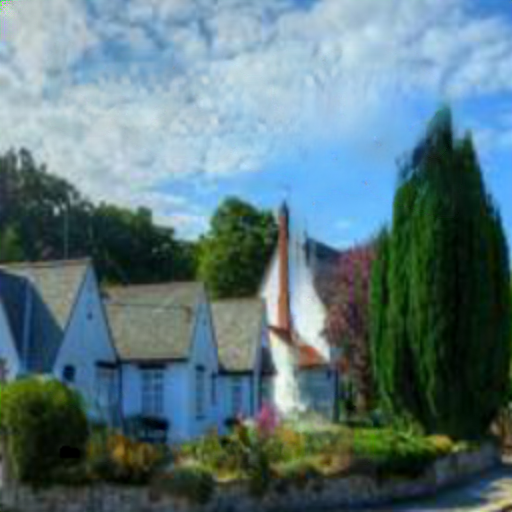}  &
		\includegraphics[width=.135\textwidth]{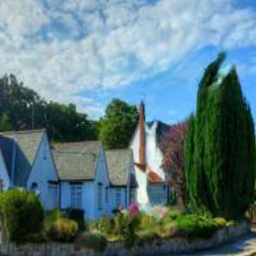}  &
		\includegraphics[width=.135\textwidth]{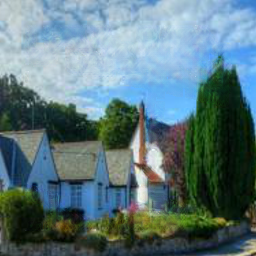}  &
		\includegraphics[width=.135\textwidth]{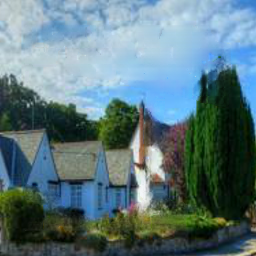}  &
		\includegraphics[width=.135\textwidth]{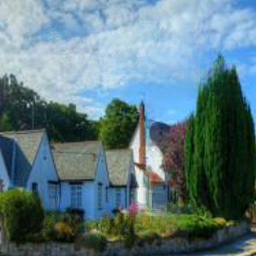}  \\
		
		\includegraphics[width=.135\textwidth]{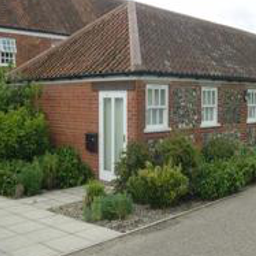}  &
		\includegraphics[width=.135\textwidth]{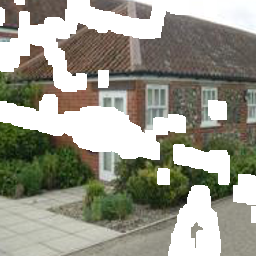}  &
		\includegraphics[width=.135\textwidth]{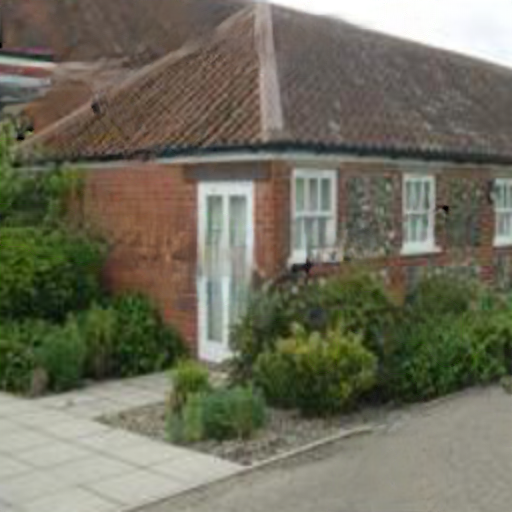}  &
		\includegraphics[width=.135\textwidth]{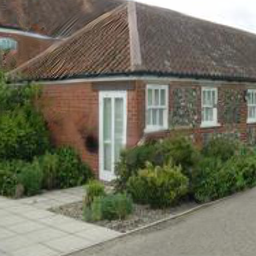}  &
		\includegraphics[width=.135\textwidth]{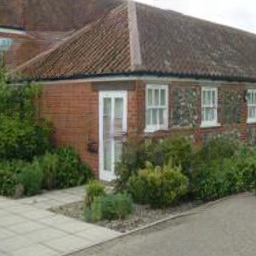}  &
		\includegraphics[width=.135\textwidth]{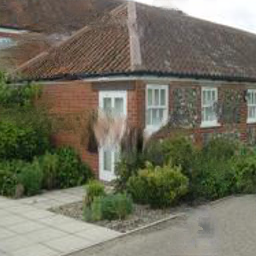}  &
		\includegraphics[width=.135\textwidth]{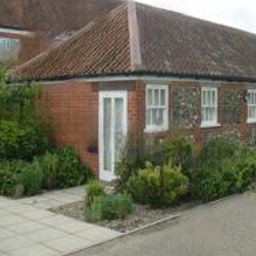}  \\
		
		\scriptsize{(a) Original} & \scriptsize{(b) Input} & \scriptsize{(c) PConv~\cite{partialconv2017}} & \scriptsize{(d) DFv2~\cite{yu2018free}} & \scriptsize{(e) EC~\cite{nazeri2019edgeconnect}} & \scriptsize{(f) MEDFE~\cite{liu2020rethinking}}  &\scriptsize{(g) Ours} \\
		\vspace{-2mm}
	\end{tabular}
	\caption{Qualitative comparison on Places2~\cite{zhou2017places} with PConv~\cite{partialconv2017}, DeepFillv2 (DFv2)~\cite{yu2018free}, EdgeConnect (EC)~\cite{nazeri2019edgeconnect}, and MEDFE~\cite{liu2020rethinking}.}
	\label{fig:places}
\end{figure*}
\begin{figure*}[hbt]
	\small
	\setlength{\tabcolsep}{2.0pt}
	\centering
	\begin{tabular}{ccccccc}
		\includegraphics[width=.135\textwidth]{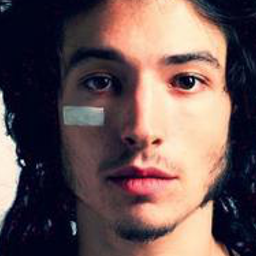}  &		\includegraphics[width=.135\textwidth]{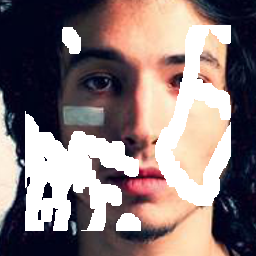}  &
		\includegraphics[width=.135\textwidth]{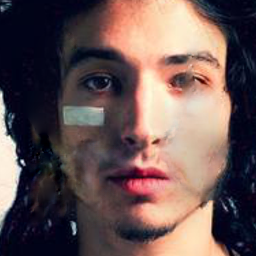}  &
		\includegraphics[width=.135\textwidth]{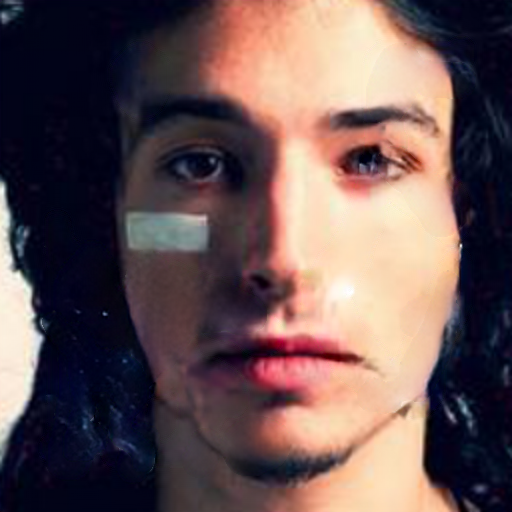}  &
		\includegraphics[width=.135\textwidth]{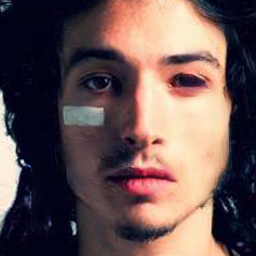}  &
		\includegraphics[width=.135\textwidth]{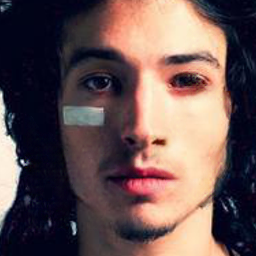}  &
		\includegraphics[width=.135\textwidth]{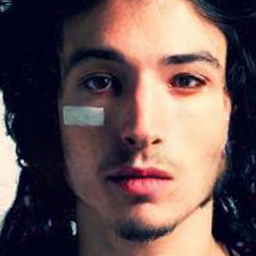}  \\
		
		\includegraphics[width=.135\textwidth]{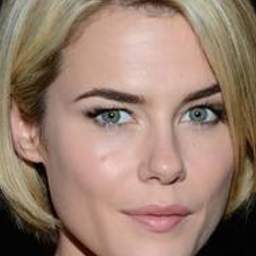}  &
		\includegraphics[width=.135\textwidth]{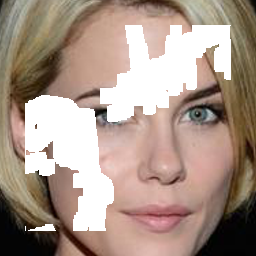}  &
		\includegraphics[width=.135\textwidth]{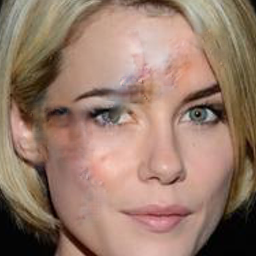}  &
		\includegraphics[width=.135\textwidth]{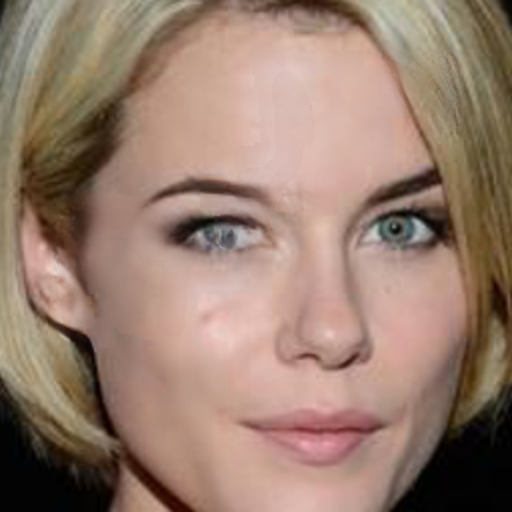}  &
		\includegraphics[width=.135\textwidth]{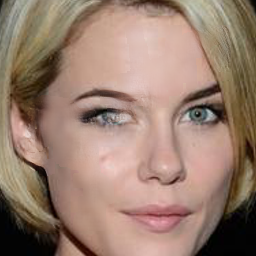}  &
		\includegraphics[width=.135\textwidth]{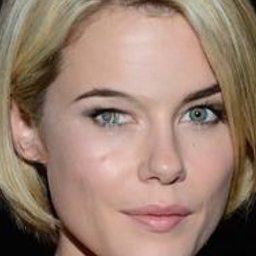}  &
		\includegraphics[width=.135\textwidth]{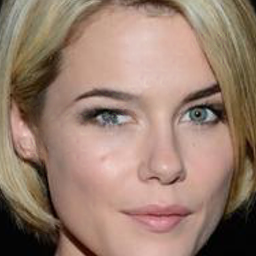}  \\
		
		\scriptsize{(a) Original} &\scriptsize{(b) Input} & \scriptsize{(c) GL~\cite{IizukaGL}} &\scriptsize{(d) PConv~\cite{partialconv2017}} & \scriptsize{(e) DFv2~\cite{yu2018free}} &\scriptsize{(f) EC~\cite{nazeri2019edgeconnect}} & \scriptsize{(g) Ours} \\
		\vspace{-2mm}
	\end{tabular}
	\caption{Qualitative comparison on CelebA-HQ256~\cite{karras2017progressive} with Global\&Local (GL)~\cite{IizukaGL}, PConv~\cite{partialconv2017}, DeepFillv2 (DFv2)~\cite{yu2018free}, and EdgeConnect (EC)~\cite{nazeri2019edgeconnect}.}
	\label{fig:celeba}
\end{figure*}

\subsection{Experimental Setting}\label{trainingdetails}
We conduct experiments on three datasets, \ie, Paris StreetView~\cite{doersch2015makes}, Places~\cite{zhou2017places}, and CelebA-HQ256~\cite{karras2017progressive}.
Using peak signal-to-noise ratio (PSNR), structural similarity index (SSIM)~\cite{wang2004image}, and LPIPS~\cite{zhang2018unreasonable} as performance measures, we compare our Edge-LBAM with Global-Local (GL)~\cite{IizukaGL}, PConv~\cite{partialconv2017}, DeepFillv2~\cite{yu2018free}, EdgeConnect~\cite{nazeri2019edgeconnect}, {and MEDFE~\cite{liu2020rethinking}}.
In the following, we introduce the datasets, competing methods, and implementation details, respectively.

{\textbf{Datasets}.} For evaluating the inpainting peformance, we adopt two datasets of scene photographs and one dataset of face images.
%
%
\begin{itemize}
  \item {\textbf{Paris StreetView}~\cite{doersch2015makes}} contains $14,900$ training images and $100$ test images collected from street views of Paris.
      We randomly select $100$ images from the original training set as our validation set, and then use the remaining $14,800$ to constitute our training set.
  \item {\textbf{Places}~\cite{zhou2017places}} contains over $2,000,000$ images from $365$ scenes.
      We randomly select $10$ categories and divide the original validation set into two equal sets to form our validation set and test set, respectively.
      While our training set contains another $50,000$ images from the same $10$ categories.
  \item {\textbf{CelebA-HQ256}~\cite{karras2017progressive}} is comprised of 30,000 human face images.
      We choose the last $200$ images to make our validation and test sets, and use the remaining $29,800$ images for training.
\end{itemize}
%





For training and testing deep image inpainting models, we generate $18,000$ irregular masks with random shapes, of which $12,000$ are from~\cite{partialconv2017}.
In our experiments, all the images are resized to make the minimum of height and width to be 350, on which $256\times256$ images are further randomly cropped.
%


{\textbf{Competing Methods}.} Using the above three datasets, our Edge-LBAM is compared with four state-of-the-art methods~\cite{IizukaGL,partialconv2017,yu2018free,nazeri2019edgeconnect}.
Global-Local (GL)~\cite{IizukaGL} applied both global and local discriminators for better inpainting semantic structures and photo-realistic details.
PConv~\cite{partialconv2017} exploited partial convolution layer for handling irregular holes.  %
DeepFillv2~\cite{yu2018free} presented gated convolutions for generalizing PConv.
EdgeConnect~\cite{nazeri2019edgeconnect} adopted a two-stage model where edge map is first predicted and then incorporated with the corrupted image for improving inpainting results.
{MEDFE~\cite{liu2020rethinking} utilized mutual encoder-decoder for filling in structures and textures, and then exploited feature equalization to make them consistent with each other.}
%
%
%
%
%
%
%
%
For evaluating inpainting performance, we adopt four measures, \ie, $\ell_1$-norm loss ($\| \overline{\mathbf{M}} \odot (\mathbf{I}_{gt} - \hat{\mathbf{I}}) \|_1 / \|\overline{\mathbf{M}} \odot \mathbf{I}_{gt}\|_1 \times 100\%$), PSNR, SSIM~\cite{wang2004image}, LPIPS~\cite{zhang2018unreasonable}.
Furthermore, user study is also carried out to assess the human perceptual quality of inpainting results by the competing methods.
%

%
%
%
%

{\textbf{Implementation Detail.}} Our approach also adopts a two-stage scheme.
First, MECNet is trained using the ADAM algorithm~\cite{AdamOptim} with the learning rate of $1 \times 10^{-4}$ and $\beta=0.1$.
Using the ground-truth edge map $\mathbf{E}_{gt}$ as the input, we then train Edge-LBAM using ADAM with the learning rate of $2.5 \times 10^{-5}$ and $\beta=0.5$.
Furthermore, MECNet and Edge-LBAM are jointly finetuned using ADAM with a lower learning rate $1 \times 10^{-5}$.
%
%
In our implementation, the mini-batch size is $8$, and flip based data augmentation is used during training.
The whole training procedure ends after $500$ epochs, where MECNet and Edge-LBAM are first trained with 400 epochs individually and then jointly finetuned with another 100 epochs.
All the experiments are conducted in the PyTorch 1.1.0 environment on a PC equipped with {Xeon E3-1230 v5 CPU}, $2$ NVIDIA GTX 2080Ti GPUs, and {32 GB RAM}.
Using Paris StreetView as an example, it takes about {120 hours} to train a Edge-LBAM model.

%
%
%
%
%
%
\begin{figure*}[hbt]
	\small
	\setlength{\tabcolsep}{2.0pt}
	\centering
	\begin{tabular}{ccccccc}
		\includegraphics[width=.135\textwidth]{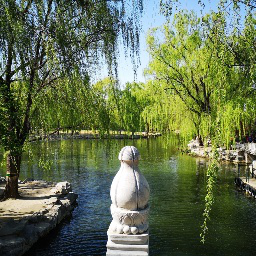}  &
		\includegraphics[width=.135\textwidth]{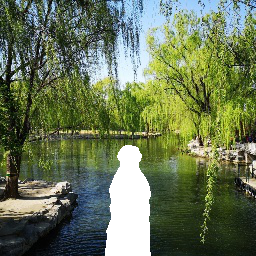}  &
		\includegraphics[width=.135\textwidth]{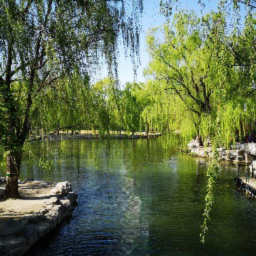}  &
		\includegraphics[width=.135\textwidth]{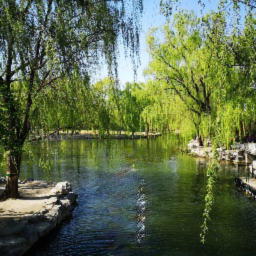}  &
		\includegraphics[width=.135\textwidth]{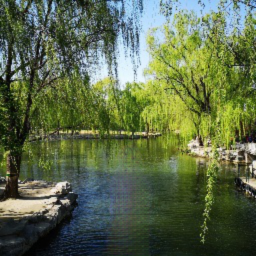}  &
		\includegraphics[width=.135\textwidth]{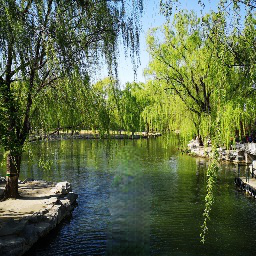}  &
		\includegraphics[width=.135\textwidth]{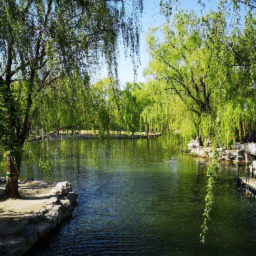}  \\
		
		\includegraphics[width=.135\textwidth]{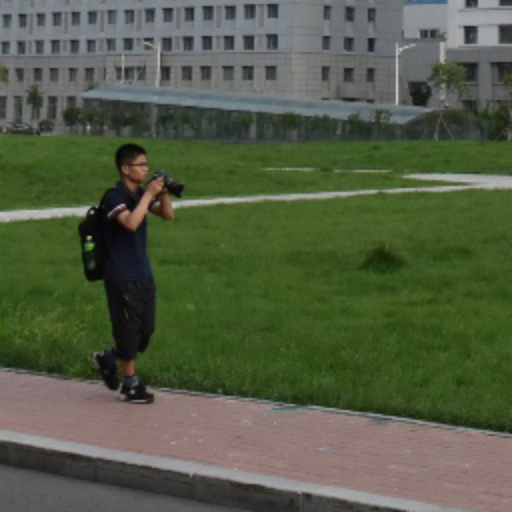}  &
		\includegraphics[width=.135\textwidth]{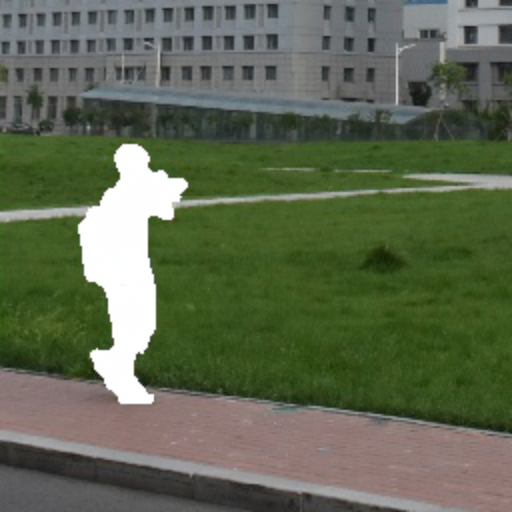}  &
		\includegraphics[width=.135\textwidth]{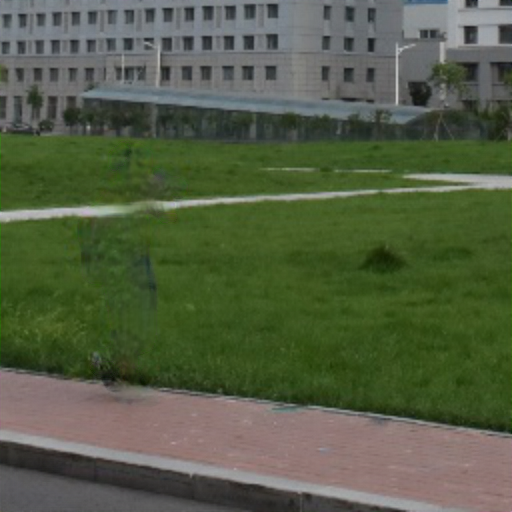}  &
		\includegraphics[width=.135\textwidth]{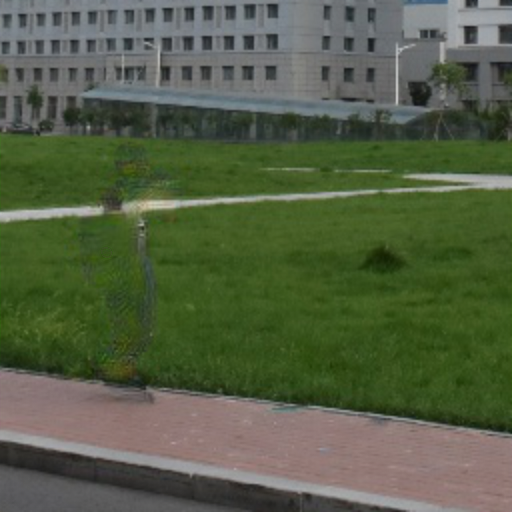}  &
		\includegraphics[width=.135\textwidth]{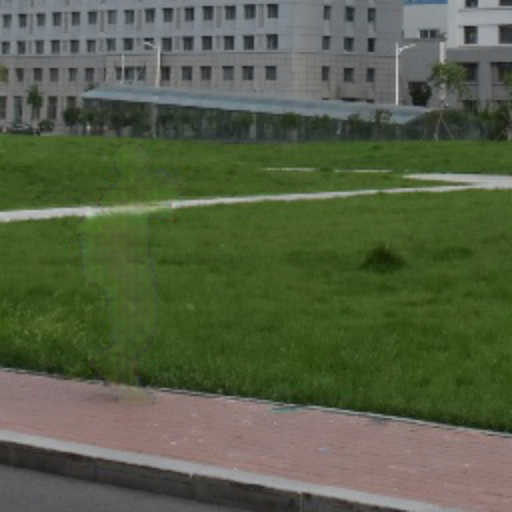}  &
		\includegraphics[width=.135\textwidth]{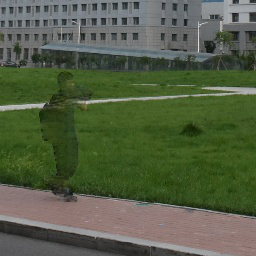}  &
		\includegraphics[width=.135\textwidth]{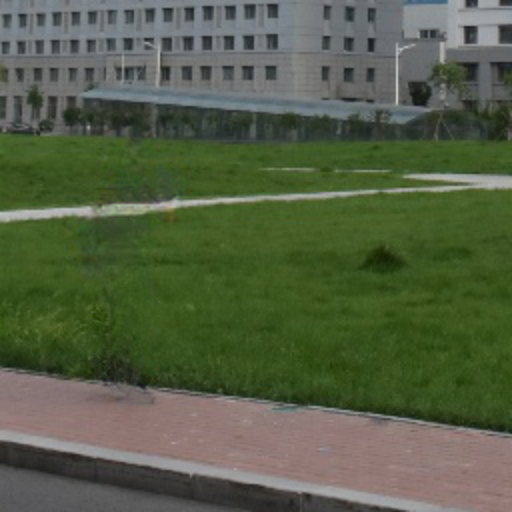}  \\
		\scriptsize{(a) Original} & \scriptsize{(b) Input} &\scriptsize{(c) PConv~\cite{partialconv2017}} &\scriptsize{(d) DFv2~\cite{yu2018free}} &\scriptsize{(e) EC~\cite{nazeri2019edgeconnect}}&\scriptsize{(f) MEDFE~\cite{liu2020rethinking}}& \scriptsize{(g) Ours} \\
		\vspace{-2mm}
	\end{tabular}
	\caption{Results on real-world images. From left to right are original image, input with objects masked (white area), Partial Convolution (PConv)~\cite{partialconv2017}, DeepFillv2 (DFv2)~\cite{yu2018free}, EdgeConnect (EC)~\cite{nazeri2019edgeconnect}, MEDFE~\cite{liu2020rethinking} and Ours.}
	\label{fig:real}
\end{figure*}

\subsection{Comparison with State-of-the-arts}\label{compare}
Quantitative and quantitative evaluation are conducted to compare our Edge-LBAM with the competing methods.
DeepFillv2~\cite{yu2018free} and EdgeConnect~\cite{nazeri2019edgeconnect} are the two methods most relevant to our Edge-LBAM.
For a fair comparison, we also implement two variants of them by respectively substituting the predicted edge map in EdgeConnect and the user sketch in DeepFillv2 with that generated by our MECNet.


{\textbf{Quantitative Evaluation}. }
Table~\ref{quantitive_results} list the mean $\ell_1$ loss, PSNR, SSIM~\cite{wang2004image}, and LPIPS~\cite{zhang2018unreasonable} values of the competing methods on Paris StreetView~\cite{doersch2015makes}, Places~\cite{zhou2017places}, and CelebA-HQ256~\cite{karras2017progressive}, respectively.
To assess the effect of hole size, we follow~\cite{partialconv2017} and report the results for filling in holes with different hole-to-image area ratios, \ie, $(0,1,0.2]$, $(0,2,0.3]$, $(0,3,0.4]$, $(0,4,0.5]$.
The official pre-trained models of PConv are not available.
Thus, its results on Places are taken from their original paper~\cite{partialconv2017}, while its results on other two datasets are based on our self-implementation and re-training.
As for other competing methods, the results are obtained using the officially released source codes and pre-trained models.
{We do not report the results of MEDFE~\cite{liu2020rethinking} on CelebA-HQ256~\cite{karras2017progressive} because MEDFE~\cite{liu2020rethinking} only provided the pretrained model for handling rectangular holes of face images.
%
}
For each test image, we use the same random mask to synthesize the same corrupted for all competing methods.

%
%
%

On the Paris StreetView dataset~\cite{doersch2015makes}, it can be seen from Table~\ref{quantitive_results} that our Edge-LBAM performs favorably against the competing methods.
In terms of PSNR, the improvement by our Edge-LBAM is more significant (\eg, $> 1$ dB) for the holes with small size.
Even for the mask with larger hole-to-image area ratio $(0,4,0.5]$, our Edge-LBAM can still achieve a PSNR gain of more than $0.30$ dB in comparison to the second best method, \ie, EdgeConnect~\cite{nazeri2019edgeconnect}.


The Places dataset~\cite{zhou2017places} is also a dataset of scene photographs.
Analogous to Paris StreetView~\cite{doersch2015makes}, we can see from Table~\ref{quantitive_results} that our Edge-LBAM performs favorably against the competing methods, and achieves higher PSNR gain for smaller holes.
Except for EdgeConnect~\cite{nazeri2019edgeconnect}, our Edge-LBAM outperforms the other competing methods by more than 0.9 dB for the holes with any size in the range $(0.1, 0.5]$.
In terms of the other quantitative measures, \eg, SSIM, LPIPS and $\ell_1$-norm loss, our Edge-LBAM is also very competitive among the competing methods.

\begin{table}[!htbp]
	\scriptsize
	\begin{center}
		\caption{User study results for various mask sizes on Paris StreetView~\cite{doersch2015makes},
			Places2~\cite{zhou2017places}, CelebA-HQ256~\cite{karras2017progressive} and Real-World Object Removal.}
		\begin{tabular}{p{1.0cm} p{1.3cm}<{\centering} p{1.3cm}<{\centering} p{1.3cm}<{\centering} p{1.3cm}<{\centering}}%
			\toprule
			
			Method & Paris & Places & CelebA-HQ  & Real-World \\
			\midrule
			GL~\cite{IizukaGL}  & 12 & 9 & 34 & 6 \\
			PConv~\cite{partialconv2017} & 41 & 53& 96 & 65      \\
			DF~\cite{yu2018free}  & 142 & 150 & 154 & 134\\
			EC~\cite{nazeri2019edgeconnect}  & 172 &167  & 180 & 158\\
			MEDFE~\cite{liu2020rethinking} & 163 &154 & --& 143 \\
			Ours& \textbf{390} & \textbf{387} & \textbf{456}  & \textbf{414}\\
			\midrule
			Total&~&~&~&\multicolumn{1}{c}{3680}\\
			\bottomrule
		\end{tabular}
		\label{user_study}
	\end{center}
\end{table}
Unlike Paris StreetView~\cite{doersch2015makes} and Places~\cite{zhou2017places}, CelebA-HQ256~\cite{karras2017progressive} is a dataset of face images.
From Table~\ref{quantitive_results}, our Edge-LBAM also performs favorably against the competing methods in terms of quantitative measures.
Considering that all face images exhibit similar semantic structures, the inpainting of face images seems to be relatively easier than the inpainting of scene photographs.
This may explain why our Edge-LBAM only achieves limited PSNR gain on CelebA-HQ256 in contrast to Paris StreetView~\cite{doersch2015makes} and Places~\cite{zhou2017places}.

%

{\textbf{Qualitative Evaluation}.} For each of the three datasets, we compare the inpainting results by our Edge-LBAM with those by the four best competing methods in terms of quantitative measures.
Figs.~\ref{fig:paris},~\ref{fig:places} and~\ref{fig:celeba} show the results by our Edge-LBAM and the competing methods on Paris StreetView~\cite{doersch2015makes}, Places~\cite{zhou2017places}, and CelebA-HQ256~\cite{karras2017progressive}, respectively.

Global\&Local~\cite{IizukaGL} is not specifically proposed for filling in irregular holes.
From Fig.~\ref{fig:paris}(c) and Fig.~\ref{fig:places}(c), color discrepancy, blurriness and unwanted texture artifacts can be observed from the inpainting results by Global\&Local~\cite{IizukaGL}.
While PConv~\cite{partialconv2017} is effective in handling irregular holes, it remains limited in recovering complex structures (see Fig.~\ref{fig:paris}(d) and Fig.~\ref{fig:places}(d)).
%
%
%
%
%
DeepFillv2~\cite{yu2018free} takes the concatenation of corrupted image and mask of holes as the input, and learns soft mask from input feature map for dynamic feature re-normalization.
Thus, DeepFillv2~\cite{yu2018free} adopts a more implicit mechanism for structure-aware inpainting, and blurriness and color artifacts can still be observed on the results of DeepFillv2 (see Fig.~\ref{fig:paris}(e) and Fig.~\ref{fig:places}(e)).
EdgeConnect~\cite{nazeri2019edgeconnect} simply concatenates the predicted edge map with corrupted image for inpainting, which may be not sufficient for fully exploiting the predicted edge map.
\begin{table}[!htbp]
	\scriptsize
	\begin{center}
		\caption{Quantitative results of DFv2+, EC+ and Edge-LBAM on Paris Street View~\cite{doersch2015makes} for the holes with any size in the range $(0.1,0.5]$.}
		\begin{tabular}{p{2.2cm}<{\centering} p{1.1cm}<{\centering} p{1.1cm}<{\centering} p{1.1cm}<{\centering} p{1.1cm}<{\centering}}%
			\toprule
			Method & PSNR &SSIM & $l_1$(\%) & LPIPS \\
			\midrule
			DFv2+ &26.58 & 0.772 & 1.76 & 0.1569 \\
			\midrule
			EC+ &27.14 & 0.787 & 1.61 & 0.1458 \\
			\midrule
			Edge-LBAM & \textbf{27.57} &\textbf{0.795} & \textbf{1.60} & \textbf{0.0794}\\
			\bottomrule
		\end{tabular}
		\label{varaints_of_dfv2}
	\end{center}
\end{table}	
From Fig.~\ref{fig:paris}(f) and Fig.~\ref{fig:places}(f), it can be seen that EdgeConnect~\cite{nazeri2019edgeconnect} may produce blurry inpainting result and sometimes even fail to recover semantic structures.
In contrast, our Edge-LBAM leverages Edge-LFAM to incorporate predicted edge map for structure-aware mask-updating, and introduces Edge-LRAM for generating complementary features to encoder feature map.
Consequently, our Edge-LBAM is very effective in generating high level semantic structures and fine-scale photo-realistic textures while preventing color artifacts and blurriness.

%
%

For assessing the generalization ability of image inpainting, we further apply the models trained on Places for object removal on real-world photographs.
Fig.~\ref{fig:real} shows the results on two images, where the masks are manually specified based on object contours.
We consider four deep inpainting models, \ie, our Edge-LBAM, PConv~\cite{partialconv2017}, DeepFillv2~\cite{yu2018free} and EdgeConnect~\cite{nazeri2019edgeconnect}.
It can be seen that our Edge-LBAM is effective in generating photo-realistic results, while color discrepancy and unwanted texture artifacts can still be observed from the results by PConv~\cite{partialconv2017}, DeepFillv2~\cite{yu2018free} and EdgeConnect~\cite{nazeri2019edgeconnect}.

%
%
%

\begin{figure*}[hbt]
	\small
	\setlength{\tabcolsep}{2.0pt}
	\centering
	\begin{tabular}{ccccccccc}
		\includegraphics[width=.104\textwidth]{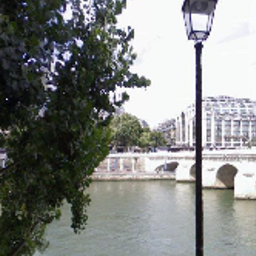}  &
		\includegraphics[width=.104\textwidth]{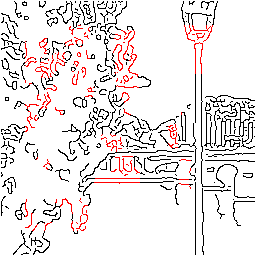}&
		\includegraphics[width=.104\textwidth]{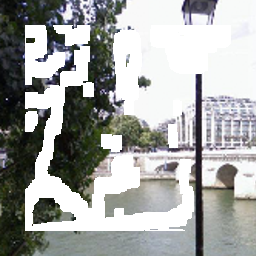}  &
		\includegraphics[width=.104\textwidth]{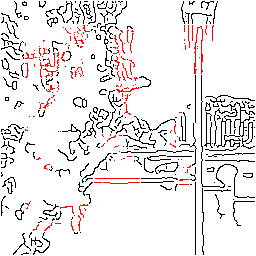} &
		\includegraphics[width=.104\textwidth]{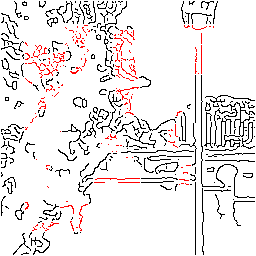}  &
		\includegraphics[width=.104\textwidth]{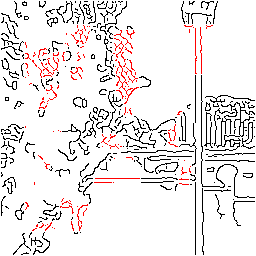}&
		\includegraphics[width=.104\textwidth]{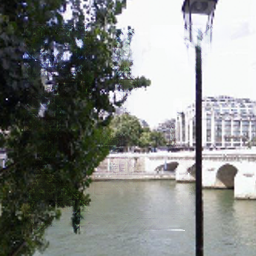}  &
		\includegraphics[width=.104\textwidth]{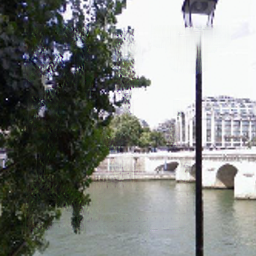}&
		\includegraphics[width=.104\textwidth]{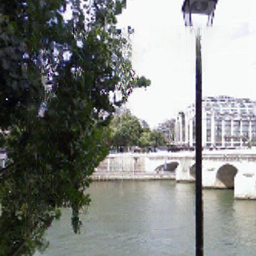} \\
		
		\includegraphics[width=.104\textwidth]{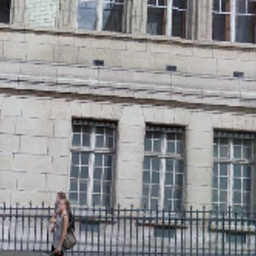}  &
		\includegraphics[width=.104\textwidth]{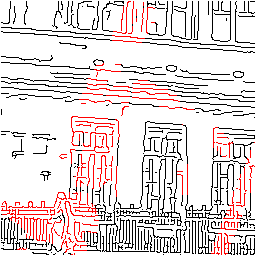}&
		\includegraphics[width=.104\textwidth]{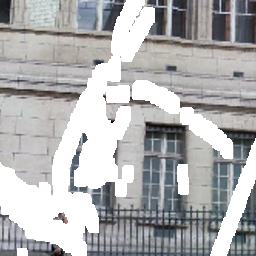}  &
		\includegraphics[width=.104\textwidth]{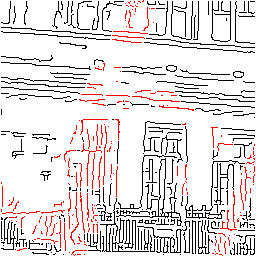} &
		\includegraphics[width=.104\textwidth]{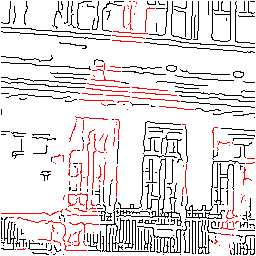}  &
		\includegraphics[width=.104\textwidth]{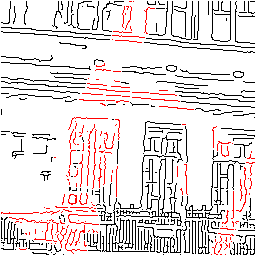}&
		\includegraphics[width=.104\textwidth]{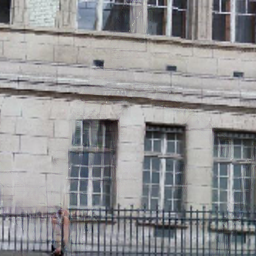}  &
		\includegraphics[width=.104\textwidth]{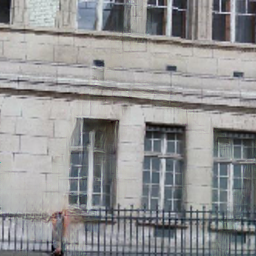}&
		\includegraphics[width=.104\textwidth]{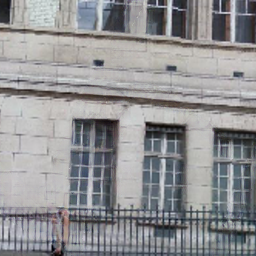} \\
		{\scriptsize{(a)}} & {\scriptsize{(b)}} & {\scriptsize{(c)}}& {\scriptsize{(d)}}&  {\scriptsize{(e)}}&{\scriptsize{(f)}} & {\scriptsize{(g)}}&  {\scriptsize{(h)}}&{\scriptsize{(i)}} \\
		\vspace{-2mm}
	\end{tabular}
	\caption{Visual quality comparison of the effect of the MECNet Variants, (a) original image, (b) Canny edge detection of (a), (c) input image with missing area, (d), (e), (f) represent the edges generated by our MECNet(S), MECNet(ML) and MECNet, (g), (h), (i) represent the inpainting results of our Edge-LBAM with the edges in (d), (e), (f). Edges drawn in black are computed (for the available regions) using Canny edge detector, whereas edges shown in red are hallucinated (for the missing regions) by the edge completion network.}
	\label{fig:single-vs-multi}
\end{figure*}

{\textbf{User Study}.} Quantitative measures, \eg, PSNR, may not well reflect human perceptual quality on images.
In contrast, quantitative study is conducted on a small number of images and is not sufficient in assessing inpainting performance.
Hence, user studies are conducted on Paris StreetView~\cite{doersch2015makes}, Places~\cite{zhou2017places}, CelebA-HQ256~\cite{karras2017progressive}, and real-world object removal for subjective visual quality evaluation.
And we consider five deep inpainting models, \ie, Global\&Local~\cite{IizukaGL}, PConv~\cite{partialconv2017}, DeepFillv2~\cite{yu2018free}, EdgeConnect~\cite{nazeri2019edgeconnect}, {MEDFE~\cite{liu2020rethinking}} and our Edge-LBAM, in the user study.



For each dataset, we randomly select $20$ test images with different hole sizes, and invite $46$ volunteers to vote for the most visually plausible inpainting result.
For each test image, the inpainting results are arranged in a random order.
Then, both the inpainting results and corrupted image are presented to the volunteer for subjective assessment.
\begin{figure}[hbt]
	\setlength{\tabcolsep}{2.0pt}
	\centering
	\begin{tabular}{ccccc}
		\includegraphics[width=.18\linewidth]{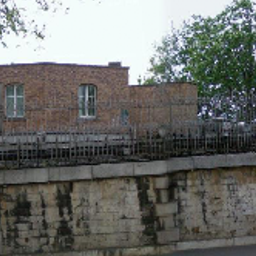}  &
		\includegraphics[width=.18\linewidth]{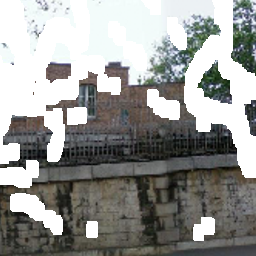}  &
		\includegraphics[width=.18\linewidth]{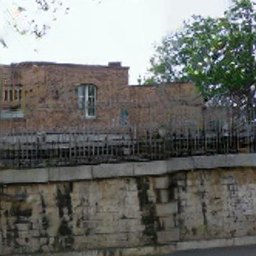}  &
		\includegraphics[width=.18\linewidth]{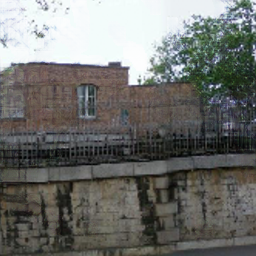}  &
		\includegraphics[width=.18\linewidth]{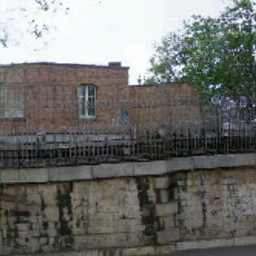} \\
		\includegraphics[width=.18\linewidth]{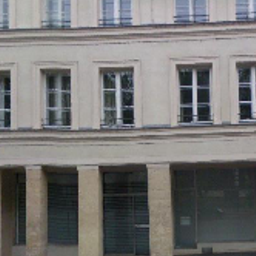}  &
		\includegraphics[width=.18\linewidth]{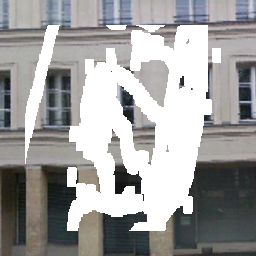}  &
		\includegraphics[width=.18\linewidth]{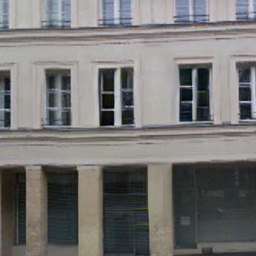}  &
		\includegraphics[width=.18\linewidth]{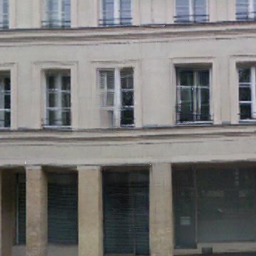}  &
		\includegraphics[width=.18\linewidth]{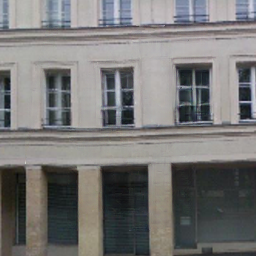}\\
		{\scriptsize{(a) Original}} & {\scriptsize{(b) input}} & {\scriptsize{(c) EC+}} &{\scriptsize{(d) DFv2+}}& {\scriptsize{(e) Ours}}\\
		\vspace{-2mm}
	\end{tabular}
	\caption{Visual quality comparison of the effect on the EC+, DeepFillv2+ (DFv2+) and our Edge-LBAM.}
	\label{fig:replace}
\end{figure}
We also provide the criteria for guiding the volunteers, which include coherency with the surrounding context, semantic structures and fine details while preventing color discrepancy, blurriness and visible artifacts.
Table~\ref{user_study} shows the user study results on the four datasets.
In general, our Edge-LBAM has about {$45\%$} chance to be selected as the winner among the four inpainting models, which is at least $25\%$ higher than the second best method, \ie, EdgeConnect~\cite{nazeri2019edgeconnect}.
It is noteworthy that our Edge-LBAM also significantly outperforms the other methods for the object removal task, clearly indicating the generalization ability of our method.

{\textbf{Comparison with Variants of DeepFillv2~\cite{yu2018free} and EdgeConnect~\cite{nazeri2019edgeconnect}}.}
EdgeConnect~\cite{nazeri2019edgeconnect} adopts an edge completion model different from our Edge-LBAM.
There are two versions of DeepFillv2~\cite{yu2018free}, where the automatic version does not take the predicted edge map as the input, and the sketch in the user-guided extension should be provided by the user.
Therefore, we implement an EdgeConnect variant (\ie, EC+) by substituting its edge completion with our MECNet, and a DeepFillv2 variant (\ie, DFv2+) by using the output of MECNet as the user sketch.
%
%
%

Using Paris StreetView~\cite{doersch2015makes}, Table~\ref{varaints_of_dfv2} and Fig.~\ref{fig:replace} compare our Edge-LBAM with EC+ and DeepFillv2+.
In terms of quantitative measures, EC+ and DeepFillv2+ respectively perform better than their counterparts, \ie, EdgeConnect~\cite{nazeri2019edgeconnect} and DeepFillv2~\cite{yu2018free}, indicating the effectiveness of MECNet.
Moreover, our Edge-LBAM outperforms EC+ and DeepFillv2+ by quantitative measures, and can generate inpainting results containing more detailed textures and being consistent with the predicted edges.
\begin{table}[!htbp]
	\scriptsize
	\begin{center}
		\caption{Quantitative results of Edge-LBAM with the predicted edges of MECNet Variants on Paris Street View.}
		\begin{tabular}{p{3.1cm}<{\centering} p{0.9cm}<{\centering} p{0.9cm}<{\centering} p{0.9cm}<{\centering} p{0.9cm}<{\centering}}%
			\toprule
			Method & PSNR &SSIM & $l_1$(\%) & LPIPS \\
			\midrule
			MECNet(S)+Edge-LBAM &27.26 & 0.792 & 1.64 & 0.0815 \\
			\midrule
			MECNet(ML)+Edge-LBAM &27.33 & 0.792 & 1.63 & 0.0819 \\
			\midrule
			MECNet+Edge-LBAM & \textbf{27.57} &\textbf{0.795} & \textbf{1.60} & \textbf{0.0794}\\
			\bottomrule
		\end{tabular}
		\label{mecnet_variants}
	\end{center}
\end{table}	
Thus, our Edge-LBAM can provide a more effective approach to leverage predicted edge map for structure-aware mask-updating and benefiting inpainting performance.

\subsection{Ablation Study}\label{ablationstudy}
Our method generally involves two stages, \ie, MECNet for edge completion, and Edge-LBAM for structure-aware inpainting.
Using Paris StreetView~\cite{doersch2015makes}, ablation studies are first carried out to assess the effect of the major components of MECNet and Edge-LBAM.
Using feature visualization, we further analyze the role of edge guidance on structure-aware mask-updating.


{\textbf{MECNet Variants}.}
Our MECNet adopts a multi-scale network, where the output of lower scale branch is upsampled and combined with the next higher-scale branch.
In our implementation, the adversarial and reconstruction losses are only added to the output of the highest scale.
Thus, we consider two MECNet variants: (i) MECNet(S) by only keeping the highest-scale branch, and (ii) MECNet(ML) by providing supervision to each scale.

Fig.~\ref{fig:single-vs-multi} illustrates the edge completion as well as image inpainting results by different MECNet variants.
While Table~\ref{mecnet_variants} lists the quantitative results of Edge-LBAM with the edge maps predicted by different MECNet variants.
The quantitative metrics are obtained by averaging the results with different hole-to-image area ratios.
Unsurprisingly, multi-scale network can produce more coherent and correct edges, and is beneficial to inpainting performance in terms of quantitative measures.
In contrast, the introduction of scale-wise supervision, \ie, MECNet(ML), has little positive effect on edge completion, and also cannot outperform MECNet by inpainting performance.
\begin{figure*}[hbt]
	\setlength{\tabcolsep}{2.0pt}
	\centering
	\begin{tabular}{cccccc}
		\includegraphics[width=.155\textwidth]{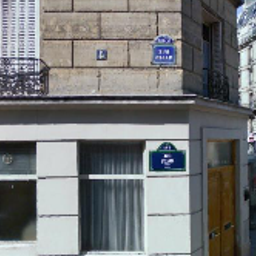} &
		\includegraphics[width=.155\textwidth]{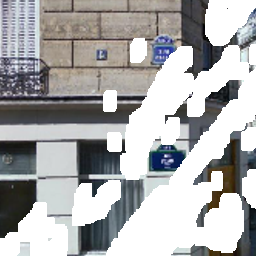} &
		\includegraphics[width=.155\textwidth]{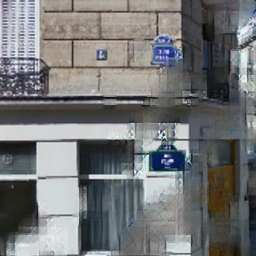} &
		\includegraphics[width=.155\textwidth]{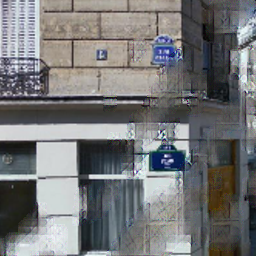} &
		\includegraphics[width=.155\textwidth]{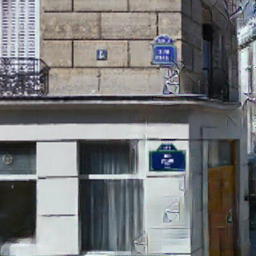} &
		\includegraphics[width=.155\textwidth]{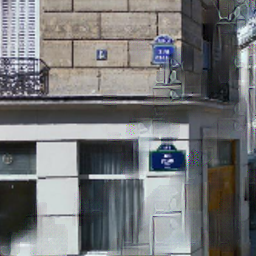}  \\
		{\scriptsize{(a) Original}} & {\scriptsize{(b) input}} &{\scriptsize{(c) BF}} & {\scriptsize{(d) BF+BR}} & {\scriptsize{(e) LFAM}} & {\scriptsize{(f) LFAM+BR}}\\
		\includegraphics[width=.155\textwidth]{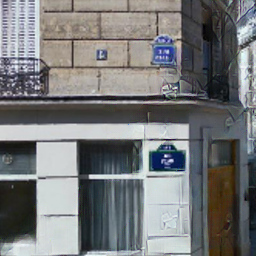} &
		\includegraphics[width=.155\textwidth]{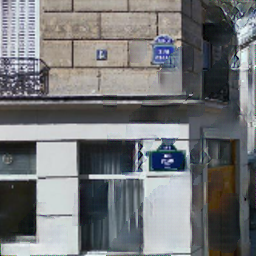} &
		\includegraphics[width=.155\textwidth]{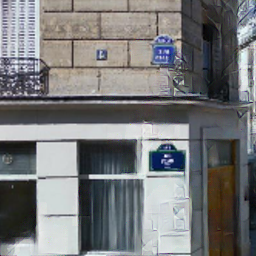} &
		\includegraphics[width=.155\textwidth]{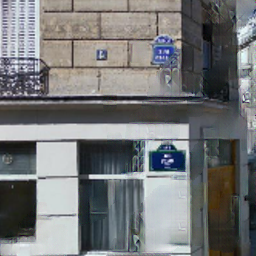} &
		\includegraphics[width=.155\textwidth]{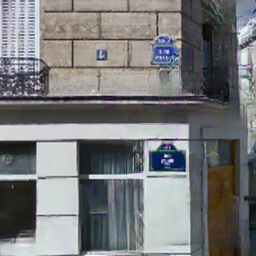} &
		\includegraphics[width=.155\textwidth]{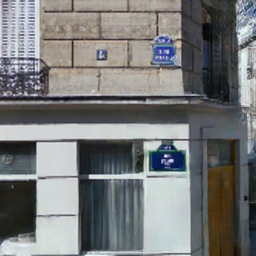} \\
		{\scriptsize{(g) LBAM}} & {\scriptsize{(h) Edge-LFAM}} & {\scriptsize{(i) LBAM(E)}} &{\scriptsize{(j) Edge-LFAM+BR}} & {\scriptsize{(k) Edge-LFAM+LRAM}} & {\scriptsize{(l) Edge-LBAM}} \\
		\vspace{-2mm}
		
	\end{tabular}
	\caption{The visual quality comparison of the effect on the LBAM and Edge-LBAM variants, (a) original, (b) input, (c)$\sim$(g) the qualitative results of LBAM variants, (h)$\sim$(l) the qualitative effect of Edge-LBAM variants.}
	\label{lbam_Edge-LBAM_variants}
\end{figure*}
\begin{table*}[!htbp]
	\scriptsize
	\begin{center}
		\caption{Quantitative results of LBAM variants and Edge-LBAM variants on Paris Street View~\cite{doersch2015makes}.}
		\begin{tabular}{p{2.0cm}<{\centering} p{2.5cm} p{2.4cm}<{\centering} p{2.5cm}<{\centering} p{2.5cm}<{\centering} p{2.5cm}<{\centering} }%
			\toprule
			&Method & 10-20\% & 20-30\% & 30-40\%  & 40-50\%\\
			\midrule
		    \multirow{10}*{\rotatebox{90}{PSNR/SSIM}}&BF & 27.31/0.869 & 24.28/0.744 & 22.38/0.685 & 21.06/0.588\\
			&BF+BR & 27.58/0.873 & 24.87/0.758 & 22.79/0.693 & 21.39/0.597 \\
		   	&LFAM  & 28.33/0.870  & 25.21/0.775 & 23.50/0.707 & 21.85/0.606\\
			&LFAM+BR  & 29.11/0.887 & 25.97/0.792 & 24.18/0.711 & 22.54/0.619\\
			&LBAM & 29.94/0.897 & 26.93/0.811 & 24.90/0.733 & 23.07/0.647 \\
			&Edge-LFAM & 29.92/0.893 & 26.71/0.803 & 24.76/0.718& 23.17/0.627\\
			&LBAM(E)  & 30.80/0.892 & 27.10/0.808 & 25.33/0.724 & 23.80/0.637 \\
			&Edge-LFAM+BR & 30.82/0.898 & 27.78/0.810 & 25.35/0.727 & 23.88/0.641 \\
			&Edge-LFAM+LRAM & 30.99/0.900 & 27.82/0.818 & 25.53/0.744 & 24.05/0.657\\
			&Edge-LBAM & \textbf{32.17}/\textbf{0.915} & \textbf{27.96}/\textbf{0.822} & \textbf{25.86}/\textbf{0.764} & \textbf{24.28}/\textbf{0.677}\\
			\midrule
			\multirow{10}*{\rotatebox{90}{$l_1$(\%)/LPIPS}} & BF & 1.17/0.0755 & 2.23/0.1523 & 3.15/0.1998 & 4.06/0.2703\\
			&BF+BR & 1.09/0.0697 & 2.08/0.1315 & 2.98/0.1776 & 3.83/0.2432 \\
			&LFAM  & 0.97/0.0519  & 1.94/0.1112 & 2.73/0.1526 & 3.46/0.1923\\
			&LFAM+BR  & 0.93/0.0462 & 1.79/0.0906 & 2.42/0.1208 & 3.26/0.1675 \\
			&LBAM & 0.72/0.0354 & 1.58/0.0735 & 2.24/0.0998 & 3.08/0.1386 \\
			&Edge-LFAM & 0.74/0.0426 & 1.70/0.0839 & 2.37/0.1167 & 3.14/0.1648 \\
			&LBAM(E)  & 0.71/0.0403 & 1.56/0.0814 & 2.19/0.1135 & 3.08/0.1629 \\
			&Edge-LFAM+BR & 0.70/0.0381 & 1.52/0.0776 & 2.12/0.1103 & 2.95/0.1592 \\
			&Edge-LFAM+LRAM  & 0.64/0.0326 & 1.35/0.0702 & 1.98/0.1004 & 2.79/0.1539\\
			&Edge-LBAM & \textbf{0.60}/\textbf{0.0235} & \textbf{1.33}/\textbf{0.0647} & \textbf{1.88}/\textbf{0.0941} & \textbf{2.59}/\textbf{0.1490} \\
			\bottomrule
		\end{tabular}
		\label{Edge-LBAM_variants}
	\end{center}
\end{table*}	
Thus, supervision loss is only added to the output of the highest scale for our default MECNet.

{\textbf{LBAM Variants}.} Five variants of LBAM are considered to assess the effect of LFAM and LRAM on inpainting result.
To begin with, we introduce a baseline method \textbf{BF} by removing LRAM, replacing the learnable $\mathbf{k}_{m}$ with $\mathbf{k}_{{1}/{9}}$, and applying the activation functions $f_A(\cdot)$ and $f_M(\cdot)$ for feature re-normalization and mask-updating.
Gradually, another baseline \textbf{BF+BR} can be provided by further substituting LRAM with $\mathbf{k}_{{1}/{9}}$, $f_A(\cdot)$ and $f_M(\cdot)$.
Analogously, we present three other LBAM variants, \ie, \textbf{LFAM}, \textbf{LFAM+BR}, \textbf{LBAM}.

%
%
Table~\ref{Edge-LBAM_variants} lists the quantitative results of LBAM variants on Paris StreetView~\cite{doersch2015makes}, and Fig.~\ref{lbam_Edge-LBAM_variants} shows the image inpainting results.
It can be seen that both LFAM and LRAM contribute to the performance gain of LBAM against the baseline methods {BF} and {BF+BR}.
Moreover, the contribution of LRAM can be more significant for handling larger holes.
From Fig.~\ref{lbam_Edge-LBAM_variants}(c)(d), the baseline methods {BF} and {BF+BR} suffer from serious blurriness, visual artifacts and color discrepancy.
LFAM can significantly reduce the visual artifacts but color discrepancy while blurriness are still visible.
By incorporating LFAM and LRAM, LBAM can further prevent color discrepancy and blurriness, but some visual artifacts remain due to the missing of structural guidance.

{\textbf{Edge-LBAM Variants}.} Edge-LBAM is proposed to incorporate predicted edge map for improving LBAM.
In particular, we implement four Edge-LBAM variants, \ie, \textbf{Edge-LFAM}, \textbf{Edge-LFAM+BR}, \textbf{Edge-LFAM+LRAM}, and \textbf{Edge-LBAM}.
Besides, we present another variant \textbf{LBAM(E)}, which extends LBAM by concatenating the predicted edge map with $\mathbf{M}$ and $\overline{\mathbf{M}}$ as the input to LFAM and LRAM, respectively.

Table~\ref{Edge-LBAM_variants} lists the quantitative results and Fig.~\ref{lbam_Edge-LBAM_variants} shows the image inpainting results of Edge-LBAM variants.
In comparison to LBAM, the quantitative result can be significantly improved by Edge-LFAM and further improved by Edge-LRAM, clearly indicating the effectiveness of Edge-LBAM.
Moreover, all Edge-LBAM variants outperform LBAM(E).
Thus, the performance gain of Edge-LBAM should be attributed to not only the incorporation of edge map, but also our proposed edge-guided module.
It can be seen from Fig.~\ref{lbam_Edge-LBAM_variants} that deployment of reverse attention maps (\eg, LRAM and Edge-LRAM) is effective in preventing visual artifacts and color discrepancy.
LBAM(E) is limited in exploiting the predicted edges, and more visual artifacts can be observed from the inpainting result.
In comparison, our Edge-LRAM is effective in generating semantic structure and detailed textures while preventing visual artifacts and color discrepancy.

%

\begin{figure*}[hbt]
	\setlength{\tabcolsep}{2.0pt}
	\centering
	\begin{overpic}[width=1\textwidth]{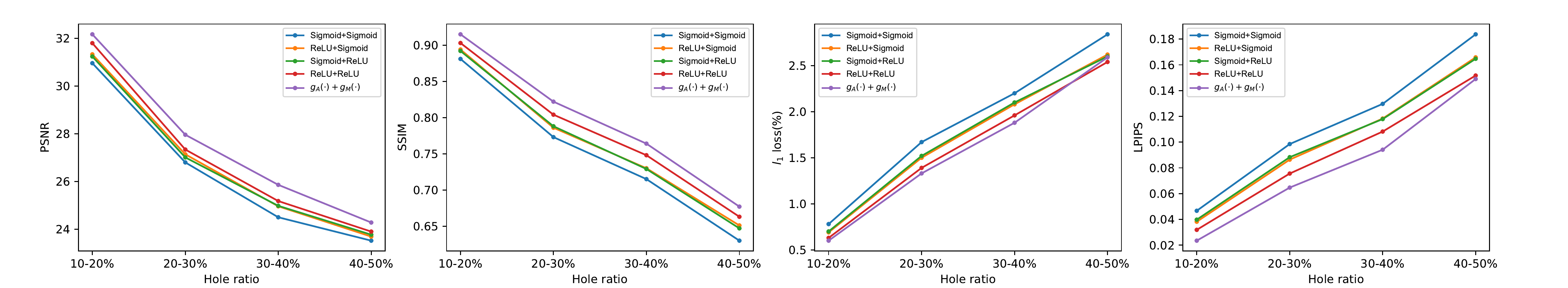}
	\put(11.5,-1.2){\scriptsize{(a) PSNR}}
	\put(35,-1.2){\scriptsize{(b) SSIM}}
	\put(58,-1.2){\scriptsize{(c) $l_1$ loss(\%)}}
	\put(82,-1.2){\scriptsize{(d) LPIPS}}
	\end{overpic}
	\vspace{-2mm}\\
	\caption{The quantitative results of Edge-LBAM variants with different activation functions, quantitative indicators from left to right are PSNR, SSIM, $l_1$ loss(\%) and LPIPS.}
	\label{activation_functions_variants}
\end{figure*}
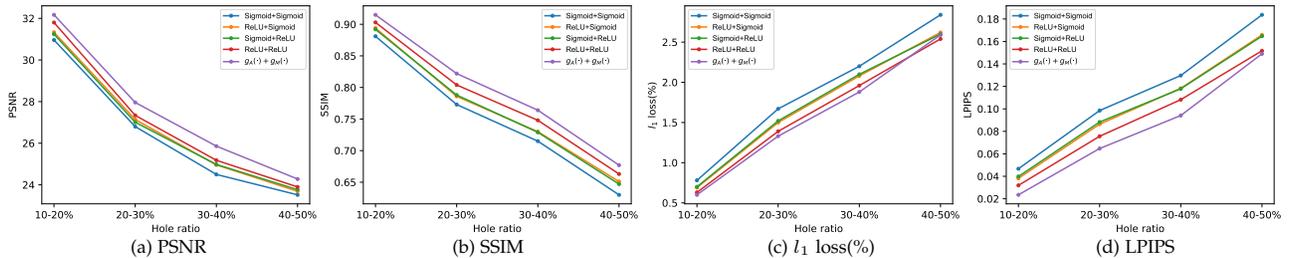
\begin{figure*}[hbt]
	\small
	\setlength{\tabcolsep}{2.0pt}
	\centering
	\begin{tabular}{cccccccc}
		\includegraphics[width=.130\textwidth]{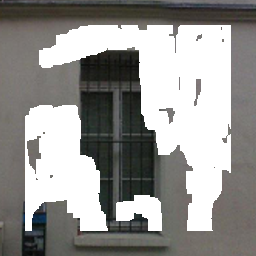}  &
		\includegraphics[width=.130\textwidth]{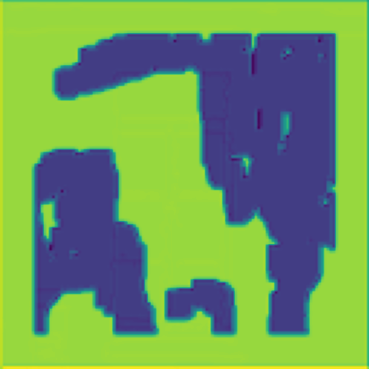}  &
		\includegraphics[width=.130\textwidth]{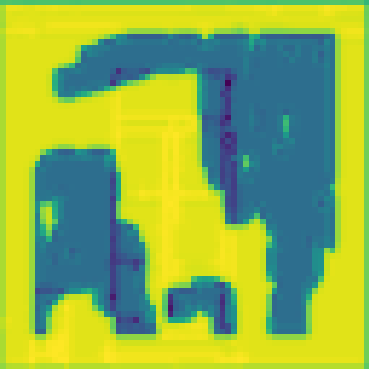}  &
		\includegraphics[width=.130\textwidth]{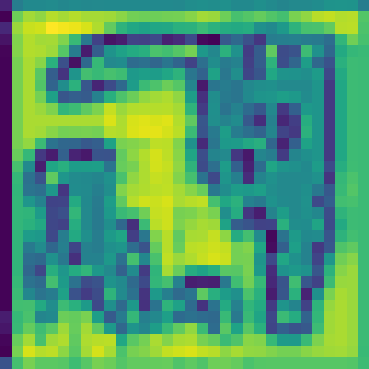}  &
		\includegraphics[width=.130\textwidth]{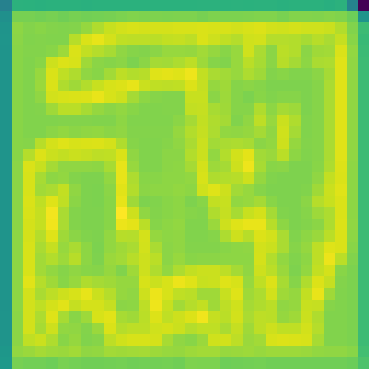}  &
		\includegraphics[width=.130\textwidth]{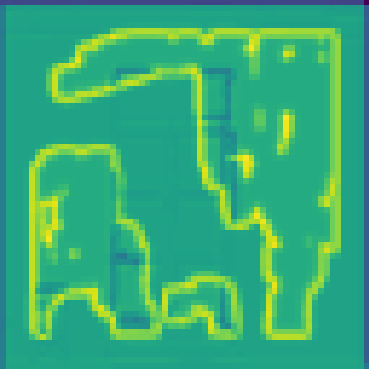}  &
		\includegraphics[width=.130\textwidth]{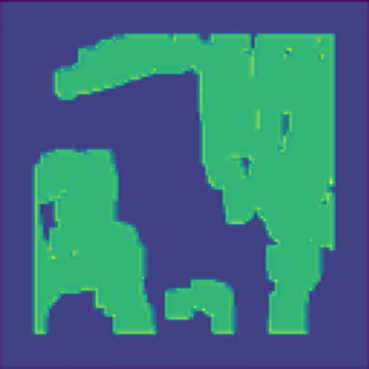} &
		\includegraphics[height=.130\textwidth]{paris/colorbar}\\
		\scriptsize{(a)} & \scriptsize{(b)} & \scriptsize{(c)} &\scriptsize{(d)} & \scriptsize{(e)} & \scriptsize{(f)} & \scriptsize{(g)} & \\
		\includegraphics[width=.130\textwidth]{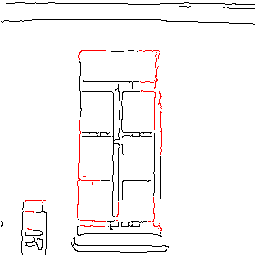}  &
		\includegraphics[width=.130\textwidth]{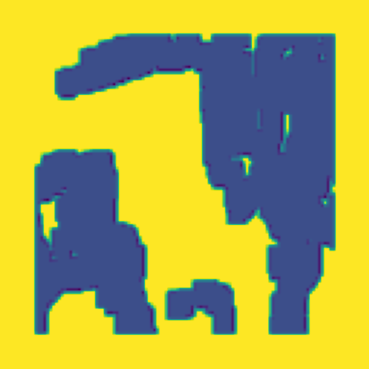}  &
		\includegraphics[width=.130\textwidth]{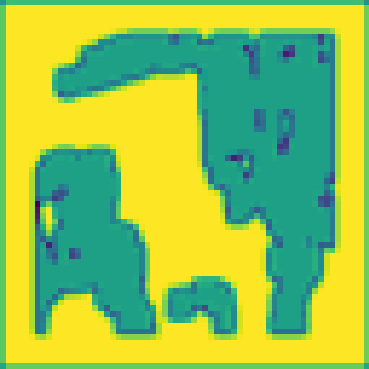}  &
		\includegraphics[width=.130\textwidth]{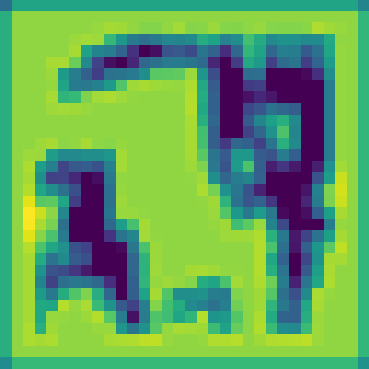}  &
		\includegraphics[width=.130\textwidth]{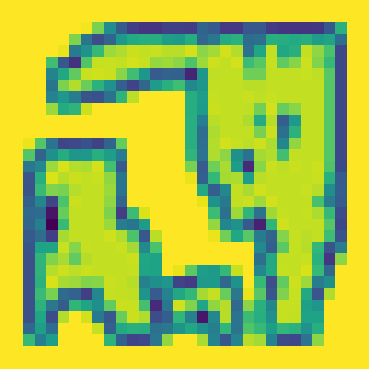}  &
		\includegraphics[width=.130\textwidth]{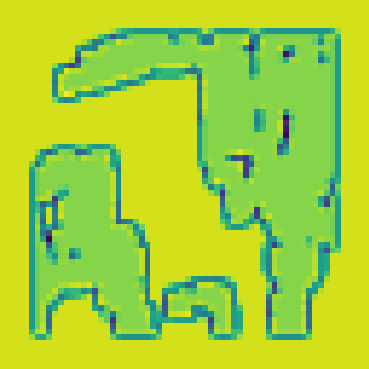}  &
		\includegraphics[width=.130\textwidth]{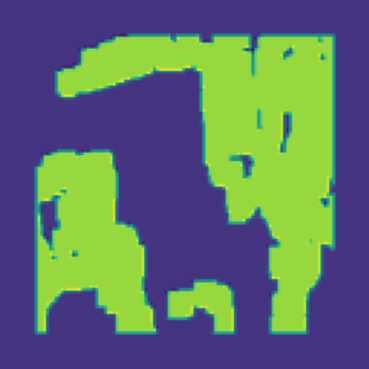} &
		\includegraphics[height=.130\textwidth]{paris/colorbar}\\
		\scriptsize{(h)} & \scriptsize{(i)} & \scriptsize{(j)} &\scriptsize{(k)} & \scriptsize{(l)} & \scriptsize{(m)} & \scriptsize{(n)} & \\
		\vspace{-2mm}
	\end{tabular}
	
	\caption{Visual comparision of updated mask maps for different layers of LBAM and Edge-LBAM. (a) input image, (b)(c)(d) forward mask maps for the first three (\ie, 1, 2, and 3) layers, (e)(f)(g) reverse mask maps from the last three (\ie, 11, 12, and 13) layers of Edge-LBAM, (h) Estimated edge map for the input in (a), (i)$\sim$(n) are the corresponding mask maps of LBAM.}
	\label{mask-updating}
\end{figure*}
{\textbf{Activation Functions for Feature Re-normalization and Mask-updating}.}
In our Edge-LBAM, the activation functions $g_A(\cdot)$ and $g_M(\cdot)$ are suggested for feature re-normalization and mask-updating.
Naturally, one may ask whether existing activation functions \eg, Sigmoid and ReLU, will work well for Edge-LBAM.
By substituting $g_A(\cdot)$ and $g_M(\cdot)$ with sigmoid or ReLU, we have four variants of Edge-LBAM, \ie, Sigmoid+Sigmoid, Sigmoid+ReLU, ReLU+Sigmoid, ReLU+ReLU.
And the original Edge-LBAM is also renamed as $g_A(\cdot)$+$g_M(\cdot)$.
Fig.~\ref{activation_functions_variants} {shows} the quantitative results of Edge-LBAM variants with different activation functions.
Unsurprisingly, $g_A(\cdot)$+$g_M(\cdot)$ significantly outperforms the other Edge-LBAM variants.  %
Thus, our $g_A(\cdot)$ and $g_M(\cdot)$ are more suitable for feature re-normalization and mask-updating than existing activation functions.
%
%
%
%
%
%
%

{\textbf{Visualization of Mask-updating}.} In comparison to PConv, our LFAM can generate continuous mask for guiding feature re-normalization, while LRAM is introduced to make the decoder focus on filling in holes.
Moreover, Edge-LFAM is further presented for structure-ware mask-updating, and Edge-LRAM is suggested to produce mask for generating complementary feature to encoder.
Thus, we visualize the masks respectively from the first three layers of the encoder, as well as the last three layers of the decoder.
For visualizing the masks in the intermediate layers, we first normalize the output mask in each layer to the range $[0, 1]$.
Then, max-pooling across channels is used to generate the visualization result.

%

%

Fig.~\ref{mask-updating} shows the visualization results of LBAM and Edge-LBAM.
For LBAM, one can see that the holes of the encoder gradually shrink along with the increase of convolutional layer.
In contrast, the mask values for known regions are nearly zeros for the last layer of decoder, and gradually become non-zeros for the preceding convolutional layers.
Moreover, for the decoder the mask values for missing regions are much higher than those for known regions, indicating that LRAM makes the decoder concentrate more on filling in holes.

As for Edge-LBAM, it can be seen that the mask-updating is structure-aware for both the encoder and decoder.
In the encoder, the holes first shrink in an omni-direction manner.
When it meets the predicted edges, Edge-LFAM begins to shrink the holes only along the direction of predicted edges for filling in the missing structural regions.
In the decoder, the mask for the last layer concentrates on holes.
For the preceding decoder layers, it begins to emphasize the predicted edges and hole boundaries, indicating that our Edge-LRAM is effective to make the decoder generate complementary feature to encoder.

%
%
%
%

%
%

\section{Conclusion}\label{sec:conclusion}
This paper presented a edge-guided learnable bidirectional attention map (Edge-LBAM) for image inpainting of irregular holes.
Our proposed method adopts a two-stage scheme by exploiting edge map predicted by multi-scale edge completion network (MCENet) for structure-aware inpainting.
To overcome the limitation of PConv, learnable forward attention map (LFAM) is suggested for learning feature re-normalization and mask-updating.
And learnable reverse attention map (LRAM) is deployed in the decoder for improving network training and inpainting performance.
Taking the predicted edge map into account, we leverage Edge-LFAM to make mask-updating and filling-in order to be structure-aware, which is incorporated with Edge-LRAM to constitute our Edge-LBAM.
Experiments show that our Edge-LBAM performs favorably against the state-of-the-art methods, and is effective in generating coherent structures and detailed textures while preventing color discrepancy and blurriness.
In future work, we will further extend Edge-LBAM by incorporating with optical flow for structure-aware and temporally consistent video inpainting.

%
%
%

\vspace{-.1in}
\ifCLASSOPTIONcompsoc
\section*{Acknowledgment}
\else
\section*{Acknowledgment}
\fi
This work is supported in part by National Natural Science Foundation of China under Grant No. U19A2073.

%

\ifCLASSOPTIONcaptionsoff
\newpage
\fi
\vspace{-.1in}
\bibliographystyle{IEEEtran}
\bibliography{IEEEabrv,mybib2}

\end{document}